\documentclass{article}

\usepackage{iclr2027_conference,times}
\iclrfinalcopy
\usepackage{amsmath,amsfonts,bm}

\def\eqref#1{equation~\ref{#1}}

\def\1{\bm{1}}

\DeclareMathAlphabet{\mathsfit}{\encodingdefault}{\sfdefault}{m}{sl}
\SetMathAlphabet{\mathsfit}{bold}{\encodingdefault}{\sfdefault}{bx}{n}

\usepackage{enumitem}
\usepackage{graphicx}
\usepackage{booktabs}
\usepackage{float}
\usepackage{placeins}
\usepackage{hyperref}
\usepackage[all]{hypcap}
\usepackage{comment} 
\usepackage{booktabs} 
\usepackage{booktabs, makecell}

\usepackage[utf8]{inputenc} 
\usepackage[T1]{fontenc}    
\usepackage{hyperref}       
\usepackage{url}            
\usepackage{booktabs}       
\usepackage{amsfonts}       
\usepackage{nicefrac}       
\usepackage{microtype}      
\usepackage{xcolor}         
\usepackage{multicol}
\usepackage{multirow}
\usepackage{caption}

\title{Brain-IT-VQA: From Brain Signals to Answers}

\author{
  \normalfont
  Roman Beliy \quad
  Matias Cosarinsky \quad
  Oliver Heinimann \quad
  Navve Wasserman \quad
  Michal Irani \\[6pt]
  {\small Weizmann Institute of Science} \\
  {\small{roman.beliy@weizmann.ac.il}} 
}

\author{
  \normalfont
  \begin{tabular}{c}
    Roman Beliy$^{1}$ \quad
    Matias Cosarinsky$^{1}$ \quad
    Oliver Heinimann$^{2*}$ \quad
    Navve Wasserman$^{1}$ \quad
    Michal Irani$^{1}$ \\[6pt]
    {\small $^{1}$Weizmann Institute of Science \quad $^{2}$EPFL} \\
    {\small \texttt{roman.beliy@weizmann.ac.il}}
  \end{tabular}
}

\begin{document}
\let\cite\citep
\vspace*{-0.5cm}
\maketitle
\lhead{}

\begingroup
\renewcommand\thefootnote{}
\footnotetext{
Project page, code, and dataset:
\url{https://matiascosarinsky.github.io/brain-it-vqa/}
}
\addtocounter{footnote}{0}
\endgroup

\renewcommand{\thefootnote}{\fnsymbol{footnote}}
\footnotetext[1]{Work done while at the Weizmann Institute of Science.}
\renewcommand{\thefootnote}{\arabic{footnote}}

\renewcommand{\headrulewidth}{0pt}

\vspace*{-0.3cm}

\begin{abstract}
\vspace{-0.1cm}
Decoding visual content from fMRI signals recorded while a person views images, and specifically answering questions about the seen images, is a long-standing challenge. While significant progress has been made in recent years in visual question answering (VQA) from fMRI, performance remains limited. Moreover, although recent models can make increasingly accurate predictions, they have rarely been used as tools for understanding the structure of visual representations in the brain.
We present \textbf{\emph{Brain-IT-VQA}}, a framework for visual question answering from fMRI. Unlike previous methods, which extract a fixed representation from the fMRI signal, our extraction is conditioned on the question itself, so what is decoded from brain activity depends on what is being asked. 
Our model substantially outperforms previous fMRI-based captioning and VQA approaches.
We further introduce \textbf{\mbox{\emph{NSD-VQA}}}, a new large-scale dataset and benchmark for visual question answering from fMRI. Unlike existing image-fMRI VQA datasets, which typically provide only a few broad and weakly controlled questions per image, \emph{NSD-VQA} provides on average 20 question-answer pairs per image, across a wide range of controlled question categories that disentangle multiple levels of visual understanding. This enables more reliable and interpretable evaluation despite limited fMRI test data.
Together, \emph{Brain-IT-VQA} and \emph{NSD-VQA} provide both a strong predictive framework and a tool for studying brain representations. Using this benchmark, we quantify which forms of visual and semantic information can be reliably decoded from fMRI responses to natural images. We further analyze the contributions of different brain regions across question types.

\end{abstract}


\FloatBarrier
\section{Introduction}
\vspace{-0.2cm}
\begin{figure}[t]
\centering
\includegraphics[width=\linewidth]{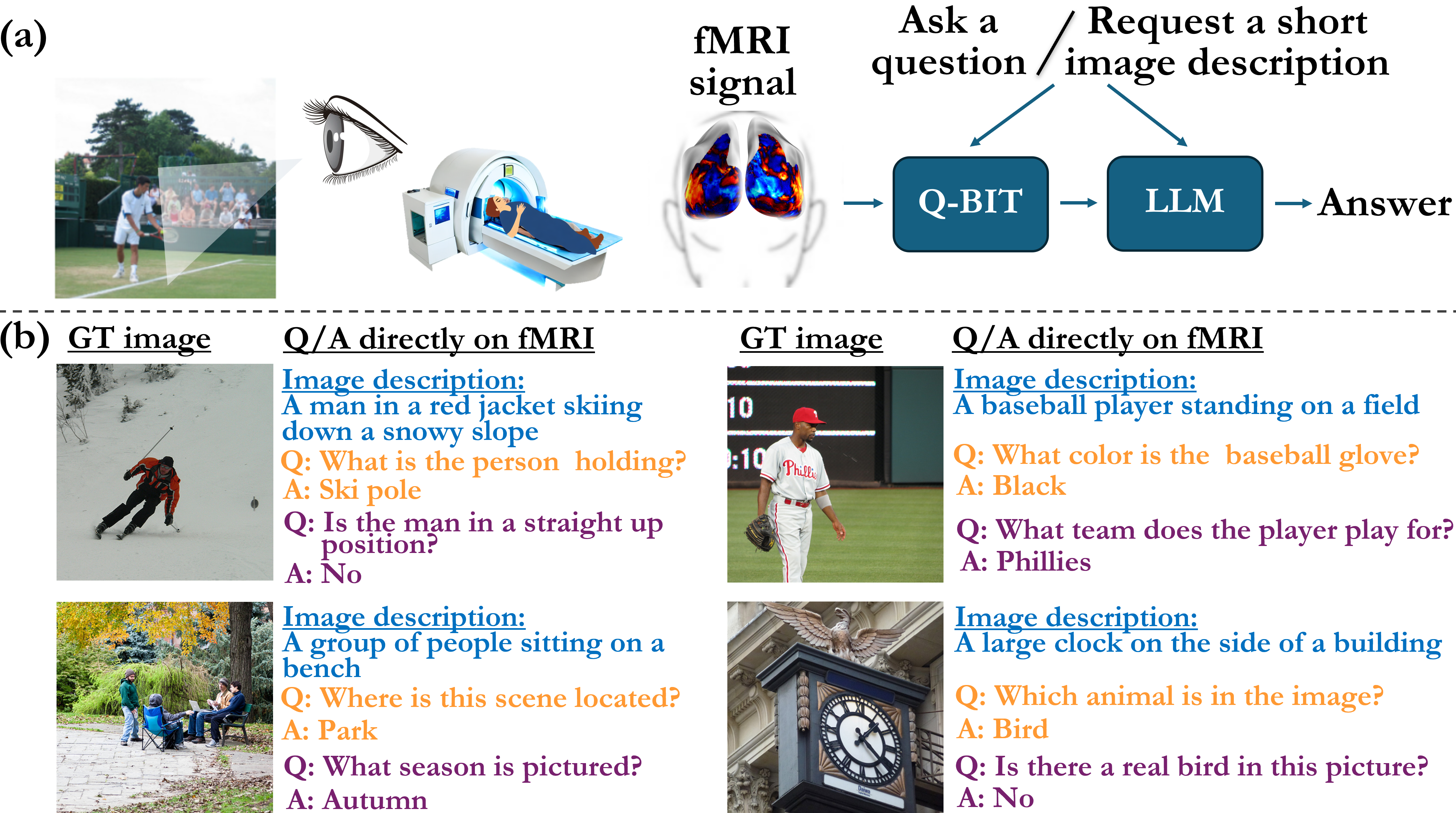}
\vspace{-0.4cm}
\caption{\textbf{fMRI-to-Language decoding: Captioning \& VQA directly
from fMRI brain activity.} {\small \it \textbf{(a)} Overview of the pipeline; fMRI signals recorded while a subject views an image are used to generate a caption or answer questions about the image. \textbf{(b)} Example captions on question-answer pairs generated by \textit{Brain-IT-VQA} directly from fMRI signals.} 
}
\label{fig:teaser}
\vspace{0.2cm}
\end{figure}

Understanding what visual information is represented in the human brain, and how different aspects of a visual scene are encoded across cortex, is a long-standing challenge in neuroscience. One approach to this question is to examine what can be decoded from functional MRI signals recorded while a person views images. Decoding visual information from fMRI can reveal information at multiple levels of abstraction, from broad semantic content, such as object and scene categories, to more specific visual properties, such as color, shape, spatial layout, and object attributes.

Recent advances in machine learning and generative models have substantially improved the ability to decode visual information from fMRI. These efforts include methods for reconstructing perceived images from brain activity 
\cite{beliy2026brainit,takagi2023high,ozcelik2023natural}
mapping brain activity to fixed visual representations 
\cite{shen2019deep,wang2024mindbridge},
and decoding fMRI into language or multimodal representations 
\cite{xia2024umbrae,wangbrainchat,mindllm2025}.
While reconstructing images or decoding specific visual properties provides an important way to study visual representations in the brain, decoding language-related representations offers a more flexible and interpretable interface for probing specific concepts and attributes. In particular, generating captions to perceived images from fMRI and visual question answering (VQA) provide direct ways to ask what information about an image can be extracted from brain activity. However, existing fMRI-based captioning and VQA models remain limited in performance, and are not necessarily aligned with the kinds of controlled aspects that are useful for neuroscientific analysis.
Since these models are typically trained and evaluated using existing vision-oriented captioning or VQA datasets, the available questions often do not target controlled neuroscientific distinctions. Together with the limited analysis of model behavior across question types and brain regions, this leaves the field still insufficiently explored.
\vspace{-0.1cm}

To address these limitations, we propose \textbf{\emph{Brain-IT-VQA}} (Fig.~\ref{fig:teaser}), a new framework for visual question answering from fMRI, together with \textbf{\emph{NSD-VQA}}, a new dataset and benchmark designed for controlled evaluation of question answering from brain activity. \textit{Brain-IT-VQA} provides a strong predictive model for answering questions about perceived images directly from fMRI, while NSD-VQA enables a more detailed analysis of which visual and semantic information can be decoded from the brain.
\textit{Brain-IT-VQA} builds on the Brain Interaction Transformer introduced in Brain-IT~\cite{beliy2026brainit}, a method for reconstructing seen images from fMRI. It
represents fMRI signals through shared functional voxel \footnote{\begingroup\fontsize{7.5pt}{9pt}\selectfont A `brain-voxel' is a small 3D cube of brain tissue, whose activity is estimated via fMRI and represented by a single scalar value. \par\endgroup} groups and integrates distributed neural information across subjects. We adapt this architecture to decode \emph{language-conditioned} representations from brain activity and combine them with a pretrained language model. Importantly, unlike prior brain-to-language methods 
the question guides our decoding, allowing different questions to extract different information from the same fMRI signal.
This allows the model to answer natural-language questions about perceived images from brain activity. \textit{Brain-IT-VQA} achieves state-of-the-art performance on fMRI-based captioning and visual question answering. 
\vspace{-0.05cm}

In addition to the model, we introduce \textbf{\emph{NSD-VQA}}, a large-scale benchmark for fine-grained evaluation of visual question answering from fMRI. 
Existing fMRI-VQA benchmarks typically contain only a small number of broad or weakly controlled questions per image, making it difficult to determine which types of information are actually decoded. 
\textbf{\emph{NSD-VQA}} addresses this by providing, on average, 20 question-answer pairs per image in the NSD dataset~\cite{allen2022massive}.
These questions are automatically generated from a wide range of 
controlled question categories that target different aspects of visual understanding, including objects, attributes, spatial relations, counting, actions, and scene-level information. This dense and structured annotation makes each measured fMRI response substantially more informative, enabling more reliable evaluation under limited test data.
\vspace{-0.05cm}

NSD-VQA allows us to move beyond a single overall VQA score and evaluate which forms of visual and semantic information can be reliably inferred from fMRI recordings of image viewing. Using this benchmark, we evaluate and analyze Brain-IT-VQA across question categories, and can further relate its predictions 
to 
brain regions. Together, Brain-IT-VQA and NSD-VQA provide a framework for using visual question answering not only as a brain-decoding task, but also as a tool for probing the organization of visual representations in the human brain.

\newpage 

\textbf{Our contributions are therefore as follows:}
\vspace{-0.3cm}

\begin{itemize}[leftmargin=*] 
    \setlength{\itemsep}{1.5pt}

    \item We introduce \textbf{\emph{Brain-IT-VQA}}, a framework for end-to-end visual question answering from fMRI, yielding state-of-the-art results.

    \item We introduce \textbf{\emph{NSD-VQA}}, a new extensive fMRI-VQA dataset specifically tailored for fMRI analysis, enabling quantitative evaluation of distinct types of visual and semantic information.

    \item We conduct a systematic empirical study of decodable information from fMRI responses to natural images, identifying which types of visual and semantic content can be reliably inferred.

    \item We provide an interpretable analysis of brain regions 
    contributions, quantifying how different 
    brain regions support distinct types of questions.
    



\end{itemize}

\vspace{-0.3cm}
\section{Related Work}
\vspace{-0.3cm}

\textbf{Vision-Language Models for VQA (on images):}
Recent advances in vision-language models (VLMs) have significantly improved performance on visual question answering and multimodal reasoning tasks. However, VLMs differ fundamentally in how they bridge visual and language modalities. One line of work, including Flamingo~\cite{alayrac2022flamingo} and the LLaVA family~\cite{liu2023llava, liu2023improvedllava}, connects visual encoders to LLMs via cross-attention gating or direct MLP projection, with the LLM fine-tuned on visual instruction data. Large proprietary systems such as GPT-4V~\cite{openai2023gpt4} and Gemini~\cite{geminiteam2023gemini} train vision and language jointly from scratch, yielding strong performance but limiting modularity and research accessibility. BLIP-2~\cite{li2023blip} introduced an alternative: a lightweight Q-Former that distils image encoder outputs into a fixed set of query token embeddings fed as soft prompts to a fully \emph{frozen} LLM. InstructBLIP~\cite{dai2023instructblip} extends this with instruction-aware feature extraction, conditioning the Q-Former on the task prompt. In our work, we draw on InstructBLIP's idea of instruction-aware extraction and adapt it to operate on brain tokens. Its Q-Former, conditioned on the question, provides the supervision target for our first training stage, guiding the brain-derived tokens toward soft prompts compatible with the frozen LLM. In the second stage, our module is then adapted end-to-end through language-model supervision, without modifying any LLM weights. 
Direct-projection models require LLM fine-tuning on image-specific features, making this substitution significantly harder. 
\vspace{-0.05cm}

\textbf{Vision-Based Brain Decoding:}
Decoding visual information from brain activity (fMRI) into perceptual and semantic representations has seen rapid progress in recent years. Early work focused on mapping fMRI signals to handcrafted or low-level visual features \cite{kay2008identifying, naselaris2009bayesian, nishimoto2011reconstructing}, followed by approaches leveraging deep neural network representations \cite{gucclu2015deep, shen2019deep, zhang2018constraint}. End-to-end methods were later introduced \cite{seeliger2018generative, st2018generative, beliy2019voxels}, followed by approaches predicting latent codes of generative models such as VAEs and GANs \cite{han2019variational, lin2019dcnn, mozafari2020reconstructing, qiao2020biggan, ren2021reconstructing}. More recent methods employ diffusion models to reconstruct images with increasing fidelity \cite{chen2023seeing, takagi2023high, ozcelik2023natural}. In parallel, there has been growing focus on leveraging shared structure across subjects to improve generalization under limited data \cite{scotti2024mindeye2, gong2025mindtuner, ferrante2024through, liu2025see, beliy2024wisdom}. 
\vspace{-0.05cm}

\textbf{Multimodal and Language-Based Decoding:}  
Recent works extend brain decoding beyond reconstruction by mapping fMRI signals to natural language or multimodal representations. Methods such as MindGPT \cite{chen2023mindgptinterpretingnoninvasivebrain}, UniBrain \cite{wang2024unibrain}, BrainCap \cite{ferrante2023braincap}, BrainChat \cite{wangbrainchat} align fMRI with visual and textual embeddings and decode language using pretrained models, typically compressing brain activity into a single shared representation. Other approaches predict intermediate stimulus representations and apply off-the-shelf visual-language models for downstream tasks \cite{xia2024umbrae}. Contrastive alignment with vision-language models has also been explored to improve captioning quality and enable region-level interpretability \cite{shen2025captioning}. 
Recent work explores end-to-end fMRI-to-text decoding with large language models across different language-based tasks \cite{mindllm2025}. 
MindLLM~\cite{mindllm2025} is the most closely related and highest performing prior method, sharing transformer-based fMRI processing and multi-subject parcellations.

Our approach differs from all previous methods in that fMRI processing is question-dependent: the information extracted from the brain signal depends on the question being asked, rather than being a fixed readout subsequently conditioned on by the LLM. Additionally, our approach differs in its processing of brain activity. Following Brain-IT, we use data-driven functional clusters rather than anatomical parcellations, which was shown to be superior for image decoding~\cite{beliy2026brainit}, and dedicated cross-attention blocks to distill task-relevant representations rather than directly prepending Brain tokens to the LLM. Brain-IT reconstructs a single image from the fMRI signal, so its output is fixed and independent of any query, whereas we condition
the extraction on the question, so that different questions draw different information from the same brain activity.

\vspace{-0.05cm}

\textbf{Datasets and Evaluation for fMRI Decoding:}
The dominant fMRI datasets for brain decoding, including NSD~\cite{allen2022massive} and BOLD5000~\cite{chang2019bold5000}, provide image--fMRI pairs, and are primarily designed for image reconstruction tasks. Evaluation in this setting often relies on pixel- and feature-level similarity metrics that capture global visual fidelity. These evaluations do not necessarily distinguish which types of visual or semantic information are recoverable from neural signals. Recent work has broadened the evaluation scope: BrainHub~\cite{xia2024umbrae} extends NSD with captioning and grounding tasks, and BrainChat~\cite{wangbrainchat} introduces fMRI question answering evaluated by classification accuracy on broad COCO VQA questions. However, neither provides controlled question categories that systematically disentangle different levels of visual understanding such as; object identity, attribute, spatial relation, or scene-level semantics. Our newly designed NSD-VQA dataset fills this gap by providing structured question types designed to probe distinct aspects of visual cognition. It enables fine-grained, interpretable evaluation of what can and cannot be decoded from fMRI responses to natural images.

\vspace{-0.1cm}
\textbf{Interpretability and Brain-Region Contributions:}  
Understanding how different brain regions contribute to decoded representations remains a central challenge. Many existing approaches compress fMRI signals into global embeddings \cite{takagi2023high, scotti2024mindeye2, wang2024mindbridge}, obscuring the contribution of individual voxels or functional regions. Some methods attempt to preserve spatial structure through voxel grouping or attention mechanisms \cite{wang2024mindbridge, huo2024neuropictor}, while others explore voxel-level or cross-subject representations ~\cite{beliy2024wisdom}. 
Recently, BrainExplore \cite{wasserman2025brainexplore} explores data-driven discovery of interpretable concepts from fMRI activity.
However, these approaches do not provide systematic analysis linking brain regions to specific decoded information. In contrast, our approach allows for direct
quantification of contributions from chosen brain regions 
across different question categories, providing insight into how brain regions support different forms of visual and semantic processing.
\vspace{-0.3cm}

\section{Method}
\vspace{-0.2cm}
\label{sec:method}
\begin{figure}[t]
\vspace{-0.5cm}

\centering
\includegraphics[width=0.95\textwidth]{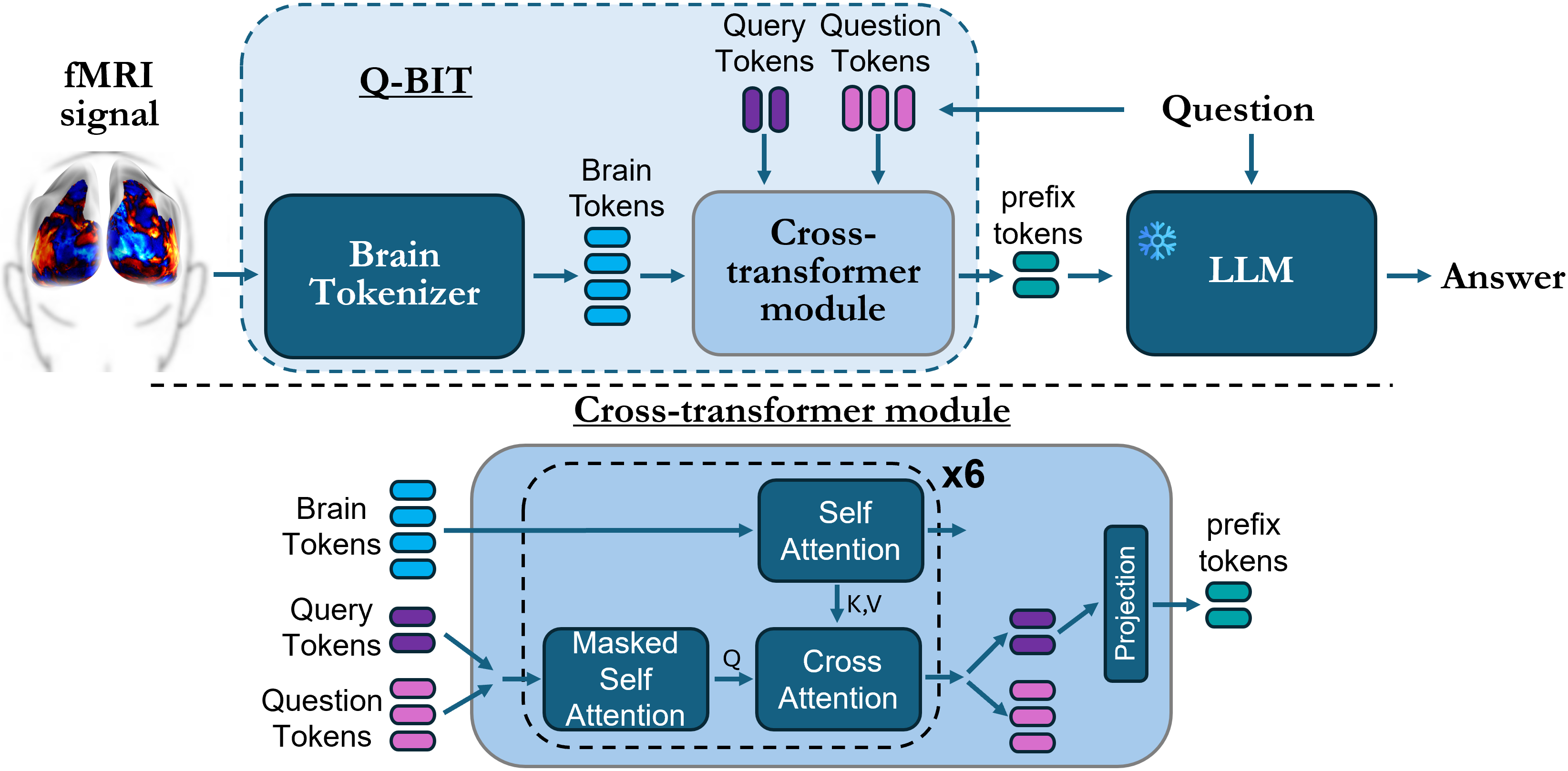}
\vspace{-0.15cm}
\caption{\textbf{Overview of the \emph{Brain-IT-VQA} architecture.}}
\label{fig:architecture}
\end{figure}


\subsection{Overview of our approach}
\vspace{-0.2cm}

We present \textbf{\emph{Brain-IT-VQA}} (Fig.~\ref{fig:teaser}), a framework for decoding fMRI brain activity into natural
language, supporting both image captioning and visual question answering (VQA).  
Given fMRI brain activity recorded while a subject views an image, the model generates either a descriptive caption or an answer to a textual query about the image 
from the fMRI signal.
Our approach extends the Brain Interaction Transformer (BIT) of ~\cite{beliy2026brainit} to predict language tokens from fMRI signals. We denote this extension as \textbf{\emph{Q-BIT}}, a question-conditioned Brain Interaction Transformer that produces soft prompts for an LLM. Q-BIT organizes voxel-level fMRI signals into clusters of \emph{functionally} similar brain-voxels 
shared across multiple subjects. Each cluster is summarized into a compact \emph{Brain Token}, and the question itself guides the attention that extracts task-relevant representations from these tokens. As such,  the information extracted from brain activity depends on what is being 
asked, thus allowing different questions to extract different information from the same fMRI response.
This 
enables direct generation of captions and answers 
from brain activity. 
For   assumptions and limitations of our method, see App.\ref{sec:limit}.

\vspace{-0.1cm}
\subsection{Model Architecture}
\vspace{-0.2cm}

Q-BIT transforms fMRI activations into a set of \emph{``Brain Tokens''} -- structured representations that summarize the activity of  clusters of functionally similar brain-voxels. Following BIT~\cite{beliy2026brainit}, the Brain Tokens interact through self-attention layers, while a cross-attention mechanism with learnable query tokens extracts task-relevant information from them. In our Brain-IT-VQA (see Fig.~\ref{fig:architecture}), the question guides this extraction.
The question is embedded into a set of \emph{question tokens}, and these are concatenated with the learnable query tokens. Together they pass through masked self-attention, where the mask prevents attention to padded positions in the question, and then serve as queries in the cross-attention over the Brain Tokens. This process is repeated across blocks, with the Brain Tokens themselves updated by self-attention in parallel, so that what is extracted from the fMRI signal depends on the question at every stage. The query token outputs are then projected into \emph{soft prompts} which, together with the textual query, condition the frozen LLM of InstructBLIP to generate answers.
Further architectural and implementation details are provided in App. \ref{sec:additional_details}.

\vspace{-0.1cm}
\subsection{Training Setup}
\vspace{-0.2cm}
Training proceeds in two stages. In the first stage (\textbf{Q-BIT Pretraining}), InstructBLIP's Q-Former serves as a teacher. Given an image and a question, it produces a question-conditioned representation, and Q-BIT is trained with an MSE loss to reproduce it from the fMRI signal alone. This aligns the brain-derived representation with the question-conditioned representation the language model expects, before any language supervision is applied.
In the second stage (\textbf{End-to-End Fine-Tuning}), Q-BIT is adapted using LoRA and trained end-to-end on either caption generation or VQA with an autoregressive cross-entropy loss over the target tokens, while the LLM remains frozen.
Further training details, including computational resources are provided in App. \ref{sec:additional_details}.

\vspace{-0.3cm}
\paragraph{Enriching the Training Data.} Both stages use fMRI recordings from the NSD dataset~\cite{allen2022massive} paired with their corresponding images. Since subject-specific data is limited, we augment the training set with additional natural images which do not have fMRI recordings, by predicting their fMRI responses using the 
Image-to-fMRI encoder
~\cite{beliy2024wisdom}. In the first stage, $\sim$120k images from the unlabeled portion of COCO are used this way. In the second stage, all COCO images which are not part of the validation or test sets are used, as fMRI responses can be predicted for any subject, regardless of whether the image was actually observed.

\begin{figure}[t]
\vspace{-0.7cm}
\centering
\includegraphics[width=0.95\textwidth]{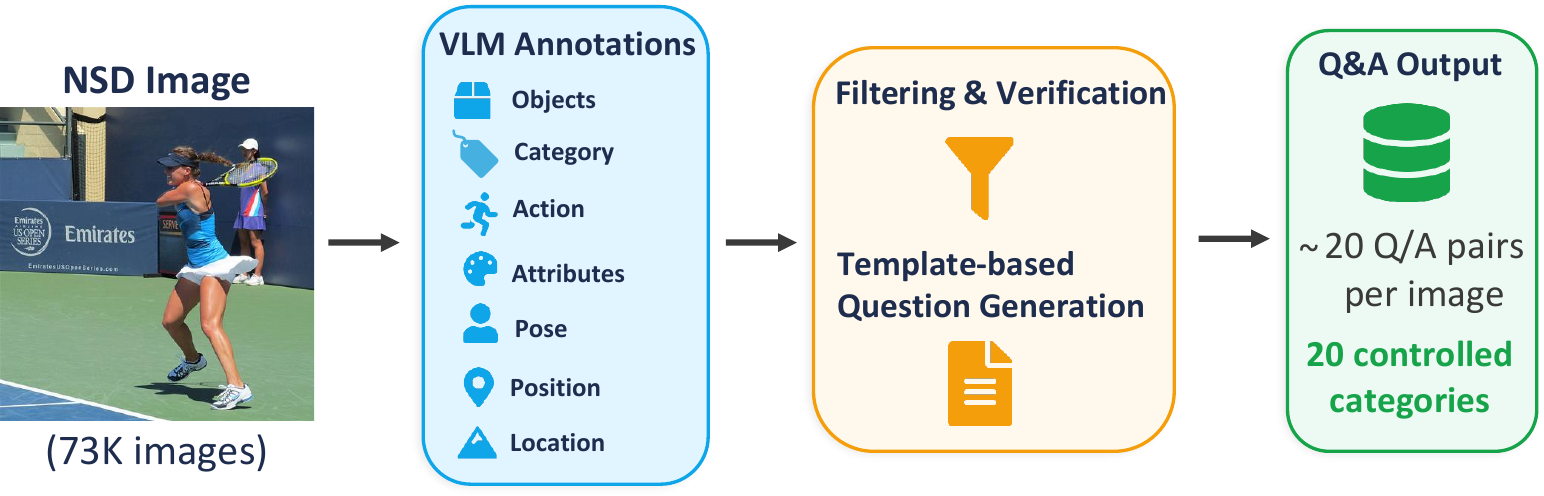}
\vspace{-0.3cm}
\caption{\textbf{\emph{NSD-VQA} Dataset construction pipeline.} {\small \it Starting from NSD images, we generate structured annotations using a VLM, followed by filtering and verification. Template-based question generation then produces multiple question–answer pairs per image across controlled question categories.}}
\label{fig:nsd-vqa}
\end{figure}
\vspace{-0.2cm}
\section{The NSD-VQA Dataset}
\label{sec:nsdvqa}
\vspace{-0.3cm}
We introduce \textbf{\emph{NSD-VQA}}, a large-scale dataset tailored for fMRI-based visual question answering. NSD-VQA is generated from the NSD dataset~\cite{allen2022massive}, which comprises 73K images with corresponding fMRI brain recordings from 8 subjects, as well as $\sim$123K unlabeled COCO images~\cite{lin2014coco}. We generate structured annotations and question-answer pairs for both sets of images, while fMRI recordings are available only for the NSD images.  
An overview of the dataset construction process is shown in Fig.~\ref{fig:nsd-vqa}. NSD-VQA is publicly available at
\url{https://huggingface.co/datasets/mcosarinsky/NSD-VQA}.

\textbf{Annotation pipeline.}
Starting from NSD images, we generate structured annotations using the vision-language model \texttt{Qwen3-VL-8B}~\cite{bai2023qwenvl}. For each image, we prompt the model to extract the most salient objects together with attributes relevant for downstream questioning. These include the following controlled categories: object identity, counts, semantic categories (e.g.\ animal, vehicle, food), color, person actions, simple interactions (e.g.\ holding), and scene-level descriptors (type and location). These categories capture distinct aspects of visual and semantic processing represented in the natural-scene images and enable controlled evaluation across question types.

\textbf{Filtering and verification.}
To improve annotation reliability, we perform a verification step focusing on object counts and presence. Counts are estimated using both \texttt{Qwen3-VL-8B} and \texttt{Gemma-4-31B-it}~\cite{gemma}, and retained only when both models agree; otherwise, the corresponding annotations are discarded. We additionally verify consistent object presence by ensuring that predicted counts are non-zero across both model predictions. Finally, a post-processing step using a BGE text encoder~\cite{bge} removes semantic redundancies by merging labels with high embedding similarity into a unified vocabulary (e.g.\ merging \emph{laptop} and \emph{notebook} into \emph{computer}). We additionally assess the quality of the generated question-answer pairs through an independent human evaluation \ref{sec:human_evaluation}. 

\textbf{Question \& Answer generation.}
From the structured annotations, we generate VQA-style question-answer pairs using templates aligned with the annotated attributes. The dataset covers object-level properties (presence, counting, color), and scene understanding (indoor/outdoor and location), as well as category-specific queries (e.g.\ animals, vehicles) and human-centric attributes (actions and interactions). Question types are instantiated conditionally based on the presence of relevant annotations (e.g.\ animal, person, or object categories), ensuring that questions are grounded in visible content. For example, an annotation specifying \textit{carrot} under the food category with an associated count of 2 would generate ``What food is in the image?'' and ``How many carrots are in the image?'', with answers ``carrot'' and ``2'', respectively.
To ensure meaningful evaluation, we augment the dataset with targeted negative examples, particularly for presence and counting questions, and enforce minimum support constraints by discarding question types with fewer than 50 instances. We further balance answer distributions across categories by filtering questions with highly skewed answers (single class dominating more than 70\% of instances).
Default answers are short-form (e.g.\ single words or short expressions), enabling controlled evaluation. We additionally construct a full-sentence variant, \textbf{NSD-VQA-FS}, by prompting a large-language model (\texttt{Llama-3.2-3B}~\cite{llama3}) to rewrite each question-answer pair into a full-sentence response.

\textbf{Summary.}
NSD-VQA provides a structured benchmark for evaluating how different types of visual and semantic information are represented in fMRI signals. By decomposing the task into targeted question categories grounded in explicit annotations, it enables systematic analysis of what information can be reliably inferred from brain activity.
All used prompts and additional details, including the question category design and human evaluation, are provided in App. \ref{sec:NSD_VQA_APP}.

\vspace{-0.31cm}
\section{Experiments}
\vspace{-0.15cm}

\subsection{Experimental Setup}
\vspace{-0.17cm}

\paragraph{Datasets.}
We train and evaluate on the Natural Scenes Dataset (NSD), a large-scale 7-Tesla fMRI dataset recording brain activity of 8 subjects viewing images drawn from COCO~\cite{lin2014coco}. Following standard practice, we use the 1,000 images shared across all subjects as the test set. 
We consider subjects 1, 2, 5, and 7, in line with prior NSD-based brain decoding works. 
For voxel selection, we adopt the post-processed version provided by Gifford et al.~\cite{gifford2023}, which includes $\sim$40k voxels, mainly from vision-related cortical areas. 
COCO provides image captions, which we use for captioning evaluation following the standard brain captioning benchmark~\cite{xia2024umbrae}. For VQA, we evaluate on four benchmarks that extend COCO with question-answer annotations: VQA-v2~\cite{goyal2017vqav2} provides multiple questions per image with 10 human answers each, covering general visual understanding.
FSVQA~\cite{fsvqa} extends this setting to full-sentence answers, requiring richer language generation.
Further, we evaluate on NSD-VQA, our proposed benchmark introduced in Sec.~\ref{sec:nsdvqa}, and its full-sentence variant NSD-VQA-FS.

\vspace{-0.2cm}
\paragraph{Evaluation Setup.}
For short-answer tasks (VQA-v2 \& NSD-VQA), we report accuracy following standard evaluation protocol~\cite{goyal2017vqav2,vqa-acc}. 
For captioning and full-sentence generation settings (FSVQA and NSD-VQA-FS), we report standard text generation metrics, including BLEU~\cite{bleu}, METEOR~\cite{meteor}, ROUGE-L~\cite{rouge}, CIDEr~\cite{cider}, and SPICE~\cite{spice}, which respectively measure n-gram precision, semantic similarity, recall via longest common subsequence, consensus with human references, and structured semantic content.
Following prior works \cite{wang2024mindbridge,xia2024umbrae,mindllm2025},
results on VQA-v2, FSVQA, and captioning are reported for subject 1. For NSD-VQA and NSD-VQA-FS, we average across subjects 1, 2, 5 \& 7.

\vspace{0.2cm}
\begin{table}[t]
\centering
\vspace{-0.6cm}

\caption{\textbf{Results on Brain captioning} {\small \it evaluated against COCO captions (subject 1).}}
\label{tab:captioning}
\vspace{-0.3cm}

\resizebox{\linewidth}{!}{%
\begin{tabular}{lcccccccc}
\toprule
\textbf{Method} & \textbf{BLEU-1} $\uparrow$ & \textbf{BLEU-2} $\uparrow$ & \textbf{BLEU-3} $\uparrow$ & \textbf{BLEU-4} $\uparrow$ & \textbf{METEOR} $\uparrow$ & \textbf{ROUGE} $\uparrow$ & \textbf{CIDEr} $\uparrow$ & \textbf{SPICE} $\uparrow$ \\
\midrule
SDRecon~\citep{takagi2023sdrecon} & 36.21 & 17.11 & 7.22 & 3.43 & 10.03 & 25.13 & 0.138 & 5.02 \\
OneLLM~\citep{han2024onellm} & 47.04 & 26.97 & 15.49 & 9.51 & 13.55 & 35.05 & 0.230 & 6.26 \\
UniBrain~\citep{mai2023unibrain} & -- & -- & -- & -- & 16.90 & 22.20 & -- & -- \\
BrainCap~\citep{ferrante2023braincap} & 55.96 & 36.21 & 22.70 & 14.51 & 16.68 & 40.69 & 0.413 & 9.06 \\
BrainChat~\citep{wangbrainchat} & 52.30 & 29.20 & 17.10 & 10.70 & 14.30 & 45.70 & 0.261 & -- \\
UMBRAE~\citep{xia2024umbrae} & 59.44 & 40.48 & 27.66 & 19.03 & 19.45 & 43.71 & 0.611 & 12.79 \\
UniBrain~\citep{wang2024unibrain} & 59.08 & 39.64 & 26.36 & 17.68 & 17.49 & 43.48 & 0.482 & 9.38 \\
MindLLM~\citep{mindllm2025} & 61.75 & 42.84 & 29.86 & 21.24 & 19.54 & 45.82 & 0.610 & -- \\
Brain-language fusion~\citep{bosch2025brainlanguagefusionenablesinteractive}  & 62.1 & 43.7 & 29.8 & 20.4 & -- & 46.0 & 0.659 & -- \\
\midrule
\textbf{BRAIN-IT VQA (Ours)} & \textbf{68.11} & \textbf{49.30} & \textbf{35.08} & \textbf{24.81} & \textbf{24.82} & \textbf{47.97} & \textbf{0.683} & \textbf{16.00} \\
\bottomrule
\end{tabular}%
}

\vspace{3mm}

\caption{\textbf{Results on VQA-v2 and FSVQA datasets for subject 1.}}
\label{tab:VQA}
\vspace{-0.3cm}

\small
\resizebox{\linewidth}{!}{%
\begin{tabular}{llcccccc|c}
\hline
\textbf{Dataset} & \textbf{Metric} & \textbf{OneLLM} & \textbf{UMBRAE} & \textbf{BrainChat} & \textbf{MindBridge} & \textbf{UniBrain} & \textbf{MindLLM} & \textbf{BRAIN-IT VQA (Ours)} \\
\hline

\multirow{1}{*}{\textbf{VQA-v2}} 
& Accuracy $\uparrow$  & 33.68 & 51.23 & 40.02 & 47.91 & 48.58 & 52.14  & \textbf{57.61} \\

\hline
\multirow{9}{*}{\textbf{FSVQA}} 
& VQA Acc. $\uparrow$ & 31.44 & 40.67 & 36.30 & 45.95 & 44.58 & 48.03 & \textbf{55.10} \\
& FSVQA Acc. $\uparrow$ & 21.02 & 0.00 & 30.22 & 40.97 & 37.87 & 43.00 & \textbf{52.00} \\
& BLEU-1 $\uparrow$   & 37.42 & 23.11 & 83.99 & 86.52 & 85.10 & 87.10 & \textbf{89.15} \\
& BLEU-2 $\uparrow$   & 31.72 & 5.86 & 78.50 & 82.28 & 80.01 & 83.03 & \textbf{86.14} \\
& BLEU-3 $\uparrow$   & 26.95 & 2.10 & 73.00 & 78.34 & 75.49 & 79.27 & \textbf{83.19} \\
& BLEU-4 $\uparrow$   & 22.48 & 1.04 & 69.73 & 74.35 & 70.73 & 75.50 & \textbf{80.17} \\
& METEOR $\uparrow$   & 26.35 & 8.93 & 44.76 & 48.63 & 46.89 & 49.05 & \textbf{52.40} \\
& CIDEr $\uparrow$    & 0.313 & 0.004 & 0.600 & 0.657 & 0.629 & 0.666 & \textbf{0.722} \\

\hline
\end{tabular}%
}

\end{table}

\vspace{-0.3cm}
\subsection{Results}
\label{sec:results}
\vspace{-0.15cm}

\textbf{Captioning.} 
Table~\ref{tab:captioning} shows the captioning results for COCO captioning evaluated on subject 1. Brain-IT-VQA achieves state-of-the-art performance across all captioning metrics in our full setting, outperforming all prior methods by a substantial margin. Compared to the strongest prior methods, our model improves BLEU-4 by \textbf{+3.57} and METEOR by \textbf{+5.28} over
MindLLM~\cite{mindllm2025}, indicating improved semantic fidelity in the decoded captions.


\textbf{Visual Question Answering.} 
Table~\ref{tab:VQA} shows the results for VQA for subject 1.
Our method achieves the best performance across all benchmarks, improving over the strongest baseline (MindLLM) by \textbf{+5.47} accuracy on VQA-v2, with consistent gains also observed on FSVQA across both accuracy and generative metrics. Results on NSD-VQA and its full-sentence variant NSD-VQA-FS are shown in Table~\ref{tab:nsd_vqa_only} (average across subjects). Our model outperforms the previous state-of-the-art (MindLLM) across all metrics. Improvements across both short-form and full-sentence settings demonstrate that our model captures richer and more detailed semantic information from fMRI signals. Additional quantitative results are provided in App.~\ref{sec:Add_results}, including captioning metrics, VQA performance, and per-category breakdowns, all reported per subject, as well as NSD-VQA performance by category against MindLLM and a question-only sanity check.
Qualitative examples are shown in Appendix~\ref{sec:Qualitative}.















\vspace{-0.2cm}
\subsection{Decoding Performance by Question Category}
\begin{figure}[t]
\vspace{-0.6cm}
\centering

\captionsetup{type=table}
\caption{
\textbf{NSD-VQA accuracy by category}
{\small \it reported as mean $\pm$ std across subjects 1, 2, 5, and 7.}
}
\label{tab:VQA_cat}

\vspace{-0.3cm}

\small
\begin{tabular}{l c | l c | l c}
\toprule
\textbf{Category} & Acc (\%) & \textbf{Category} & Acc (\%) & \textbf{Category} & Acc (\%) \\
\midrule

action & 58.26 $\pm$ 3.21
& electronic Y/N & 89.35 $\pm$ 3.51
& person Y/N & 93.36 $\pm$ 0.67 \\

animal & 64.56 $\pm$ 3.57
& food & 50.28 $\pm$ 6.57
& scene & 93.78 $\pm$ 0.88 \\

animal Y/N & 91.13 $\pm$ 2.29
& food Y/N & 94.42 $\pm$ 0.17
& sport Y/N & 93.46 $\pm$ 1.63 \\

appliance Y/N & 94.24 $\pm$ 2.56
& holding & 54.47 $\pm$ 4.90
& street Y/N & 87.86 $\pm$ 1.54 \\

building Y/N & 82.61 $\pm$ 6.15
& household Y/N & 90.06 $\pm$ 0.72
& structure Y/N & 88.83 $\pm$ 1.06 \\

clothing Y/N & 94.74 $\pm$ 3.04
& location & 54.82 $\pm$ 2.59
& vehicle & 67.23 $\pm$ 4.92 \\

color & 43.14 $\pm$ 0.91
& nature Y/N & 83.76 $\pm$ 0.38
& vehicle Y/N & 89.91 $\pm$ 1.86 \\

counting & 79.44 $\pm$ 1.01
& objects Y/N & 89.21 $\pm$ 2.82
& & \\

\bottomrule
\end{tabular}

\vspace{3mm}

\includegraphics[width=\linewidth]{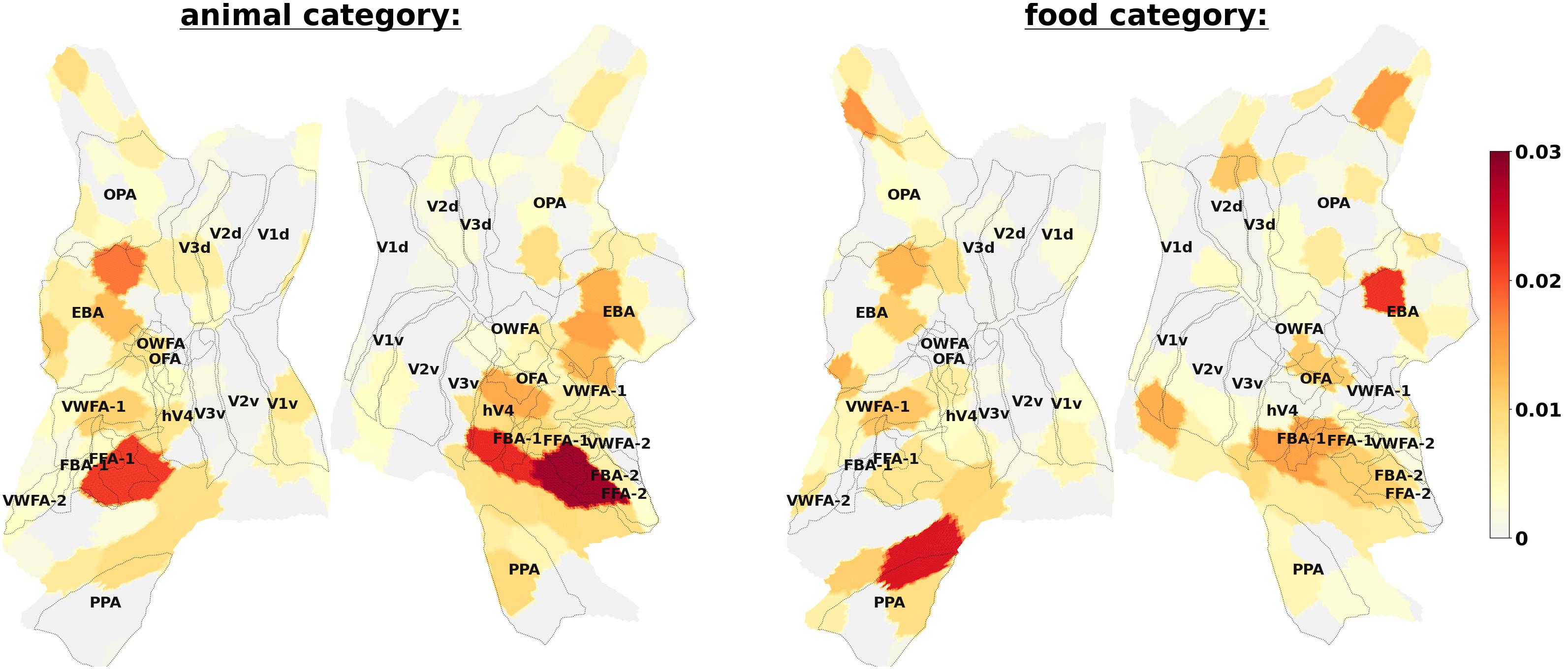}

\vspace*{-0.15cm}

\captionsetup{type=figure}
\caption{
\textbf{Visualization of Schaefer-1000 parcel marginal contributions across question categories.}
{\small \it The marginal contribution of a region refers to the change in decoding performance when that region is added to a coalition (subset) of other regions.
Different parcels vary in importance depending on the question type, highlighting how distinct brain regions support different aspects of visual and semantic processing.}
}
\label{fig:cluster_contributions}

\end{figure}

\vspace{-0.1cm}
\label{subsec:decodingperformance}
We leverage NSD-VQA to analyze which types of visual and semantic information can be reliably decoded from fMRI. By organizing questions into controlled categories, the benchmark enables fine-grained evaluation beyond aggregate accuracy. This setup allows us to examine systematic differences across object recognition, attributes, and relational reasoning. Results averaged across subjects 1, 2, 5, and 7 are shown in Table~\ref{tab:VQA_cat}, with per-subject results provided in App.~\ref{sec:additional_nsd_vqa_category}.

We observe a clear dependence on question type. Binary (Y/N) questions consistently achieve high accuracy (typically 83--95\%), reflecting the relative simplicity of binary decision tasks and indicating that fMRI signals support robust decoding of coarse object presence and categorical distinctions. 
In contrast, open-ended questions that require selecting among multiple semantic alternatives are substantially more challenging, with lower performance for categories such as \emph{color} (43.14\%), \emph{food} (50.28\%) and \emph{action} (58.26\%).


Overall, these results suggest that fMRI-based decoding preferentially captures coarse, high-level visual and categorical information, while remaining limited in resolving fine-grained semantic attributes. This pattern is consistent across question categories and highlights a gap between coarse recognition and detailed semantic discrimination. Failure-mode analyses are provided in App.~\ref{sec:failure_modes}.

\begin{table}[t]
\vspace{-0.5cm}
\centering
\caption{
\textbf{Results on NSD-VQA}. {\small \it Values are reported as mean $\pm$ std across subjects 1,2,5 and 7. NSD-VQA-FS denotes the full-sentence variant.}
}
\vspace{-0.2cm}
\small

\begin{tabular}{lccc}
\toprule
\textbf{Dataset} & \textbf{Metric} & \textbf{MindLLM} & \textbf{BRAIN-IT VQA (Ours)} \\
\midrule

\multirow{1}{*}{\textbf{NSD-VQA}} 
& Accuracy $\uparrow$  
& 73.43 $\pm$ 0.97 
& \textbf{78.23 $\pm$ 1.08} \\

\midrule

\multirow{9}{*}{\textbf{NSD-VQA-FS}} 

& FSVQA Acc. $\uparrow$
& 73.83 $\pm$ 0.66
& \textbf{78.18 $\pm$ 1.05} \\

& BLEU-1 $\uparrow$   
& 93.18 $\pm$ 0.18 
& \textbf{94.42 $\pm$ 0.29} \\

& BLEU-2 $\uparrow$   
& 91.21 $\pm$ 0.24 
& \textbf{92.80 $\pm$ 0.38} \\

& BLEU-3 $\uparrow$   
& 89.29 $\pm$ 0.30 
& \textbf{91.22 $\pm$ 0.48} \\

& BLEU-4 $\uparrow$   
& 87.03 $\pm$ 0.37 
& \textbf{89.42 $\pm$ 0.58} \\

& METEOR $\uparrow$   
& 60.13 $\pm$ 0.31 
& \textbf{62.33 $\pm$ 0.55} \\

& ROUGE-L $\uparrow$   
& 92.41 $\pm$ 0.20 
& \textbf{93.74 $\pm$ 0.31} \\

& CIDEr $\uparrow$    
& 0.801 $\pm$ 0.005 
& \textbf{0.836 $\pm$ 0.008} \\

\bottomrule
\end{tabular}
\vspace{2mm}
\label{tab:nsd_vqa_only}
\end{table}

\vspace{-0.2cm}
\subsection{Ablations}
\vspace{-0.2cm}

We conduct an ablation study to evaluate the contribution of Q-BIT and of the external data augmentation (Table~\ref{tab:ablation}). We consider the following ablations: replacing Q-BIT with the original BIT architecture of Brain-IT~\cite{beliy2026brainit}, designed for image reconstruction, which extracts a single representation per fMRI signal regardless of the question (w/o Q-BIT), and removing external data augmentation from predicted fMRI responses (w/o external data). Both contribute meaningful improvements (\textbf{+1.54} and \textbf{+2.47} respectively on VQAv2 Acc.), with larger gains on FSVQA Acc. (\textbf{+6.17} and \textbf{+4.07}), where full-sentence answers are required. Additional ablations are provided in App.\ref{sec:additional_ablations},
where we show that even without external data augmentation we outperform all previous methods on VQA, and lead on 7 out of 8 captioning metrics.

\begin{table}[h]
\vspace{-0.8cm}
\centering
\small
\setlength{\tabcolsep}{6pt}
\vspace{0.4cm}
\caption{\textbf{Ablation study.} Effect of predicted-fMRI data enrichment (Ext.\ Data) and question conditioning (Q-BIT, vs.\ the question-independent BIT readout). $\checkmark$/$\times$ indicate whether each is used.}\label{tab:ablation}
\begin{tabular}{cccc}
\toprule
Ext.\ Data & Q-BIT & VQAv2 Acc. & FSVQA Acc. \\
\midrule
$\checkmark$ & $\checkmark$ & \textbf{57.61} & \textbf{52.00} \\
$\times$     & $\checkmark$ & 55.14 & 47.93 \\
$\checkmark$ & $\times$     & 56.07 & 45.83 \\
$\times$     & $\times$     & 53.75 & 39.7 \\
\bottomrule
\end{tabular}
\vspace{0.2cm}
\end{table}

\vspace{-0.2cm}

\vspace{-0.2cm}
\section{Decoding Contribution Analysis}
\vspace{-0.2cm}


\paragraph{Overview:}
We leverage Brain-IT-VQA to analyze which brain regions encode information relevant to specific types of visual and semantic understanding. We estimate the marginal contribution %
of each brain region to decoding performance across the controlled question categories of NSD-VQA.
The marginal contribution of a region refers to the change in decoding performance when that region is added to a coalition (subset) of other regions, compared to the prediction without it.
Masking any single region is likely to underestimate its contribution, as other regions can compensate due to
the distributed and redundant nature of visual representations in the brain~\cite{haxby2001distributed,norman2006beyond}.
Instead, we adopt a randomized masking approach that accounts for this redundancy by estimating contributions across many masking configurations.

\vspace{-0.3cm}

\paragraph{Technical Details:}
We estimate the marginal contribution of each brain region to each question category using a masking-based regression approach, inspired by perturbation-based attribution methods,
specifically occlusion sensitivity~\cite{zeiler2014visualizing} and local surrogate modeling over
perturbed inputs~\cite{ribeiro2016lime}. Regions are defined by the 1000 parcels of the Schaefer atlas~\cite{schaefer2018local}, of which 198 fall within our global voxel mask. At each trial, a random subset of these parcels is masked ~\ref{APP:masking}, and the model generates predictions on 200 stimuli drawn
from the NSD test set. This is repeated for 5,000 trials, yielding a dataset of masking
configurations and their corresponding per-category VQA accuracy. We then fit a ridge regression
model with the binary masking vector over parcels as input and the
per-category score as output. The resulting regression coefficients provide an estimate of each parcel's marginal contribution to decoding performance for that category, while controlling for the contributions of all other parcels.

\vspace{-0.3cm}
\paragraph{Analysis:}
Fig.~\ref{fig:cluster_contributions} shows the estimated parcel contributions
for the \emph{animal} and \emph{food} question categories for subject 1.
We observe distinct contribution patterns across the two categories, suggesting that different
types of visual and semantic information rely on partially different brain representations.
The \emph{animal} category exhibits contributions distributed across lateral and ventral
occipitotemporal cortex, peaking near FBA and FFA, with additional contributions
around EBA and OFA, consistent with the involvement of body- and form-selective regions in
recognizing animate categories.
In contrast, contributions for \emph{food} questions are more spatially concentrated in a small
number of parcels near PPA and EBA, broadly consistent with recent findings of food-selective
representations in ventral visual cortex~\cite{khoslafood}.
These results suggest that different question categories engage distinct brain representations,
and that the nature of the information, whether localized or distributed, is reflected in the
spatial structure of the contributions. This demonstrates the potential of our framework as a
tool for probing visual-semantic organization in the brain.
Additional visualization results across additional question categories, subjects, and functional
ROIs, together with an extended discussion of the contribution patterns, are provided in
App.~\ref{APP:analysis}.

\vspace{-0.2cm}
\section{Conclusion}
\vspace{-0.3cm}
We present \emph{Brain-IT-VQA}, a SotA framework for visual Captioning \& VQA 
from fMRI brain recordings, which outperforms previous methods by a large margin. Unlike previous approaches, the extraction of information from the brain signal is conditioned on the question itself, so that different questions draw different information from the same recording.
We further introduce \emph{NSD-VQA}, a new extensive automatically-curated benchmark dataset, which provides $\sim$20 question-answer pairs per image, 
across 20 controlled question categories.
This enables more reliable and interpretable evaluation of VQA from fMRI than ever before.
Moreover, using this benchmark we can analyze the contributions of different brain regions across question types. We provide initial results for attributing information content to functional brain regions via a masking-based analysis, demonstrating that different regions contribute selectively to different types of visual and semantic understanding.
Part of our future work is a systematic evaluation of these attribution results against known functional neuroimaging literature, to assess the extent to which the identified contributions align with established region-function mappings, and whether they discover new ones.



%
%

\newpage

\section*{Acknowledgments}
This research was funded by the European Union (ERC grant No. 101142115).





\bibliography{main}
\bibliographystyle{iclr2027_conference}

\clearpage
\appendix
\renewcommand{\theHfigure}{\Alph{section}.\arabic{figure}}
\renewcommand{\theHtable}{\Alph{section}.\arabic{table}}
\begin{center}
    {\LARGE \bfseries Appendix}
\end{center}

\setcounter{figure}{0}
\setcounter{table}{0}
\setcounter{equation}{0}

\renewcommand{\thefigure}{S\arabic{figure}}
\renewcommand{\thetable}{S\arabic{table}}
\renewcommand{\theequation}{S\arabic{equation}}

\section{Limitation}
\label{sec:limit}

\paragraph{fMRI Assumptions.} Following standard practice in the field, our model assumes that fMRI responses are memoryless and replicable. The memoryless assumption implies that prior stimuli do not influence the response to the current image, while replicability assumes that repeated presentations of the same image yield consistent responses. The latter is important for signal averaging, a common practice to mitigate the low SNR of fMRI. These assumptions may not hold in all settings, as they neglect effects such as representational drift over time.
\paragraph{Subject Variability.} There is significant variability in signal quality across subjects. The interpretability analysis we present is most reliable for subjects with high SNR, where voxel functionality can be estimated more accurately. For subjects with poor signal quality, the estimated contributions of brain regions may be less reliable.
\paragraph{Interpretability Analysis.} The masking-based analysis presented in this work is intended as a demonstration of the framework's potential for brain exploration rather than a definitive functional mapping. The attribution estimates depend on the quality of the underlying VQA model and the coverage of the NSD test set, and should be interpreted with appropriate caution.

\paragraph{Model versus Brain Attribution.}

Our pipeline relies not only on fMRI measurements, but also on many other components, including pretrained vision-language and language models, synthetic fMRI generated by an image-to-fMRI encoder, and engineering choices in clustering and training. Consequently, strong performance of our method may reflect both information present in the fMRI data and prior information introduced by other components of the pipeline. Our method can therefore be used to generate hypotheses about brain representations, but these hypotheses require targeted experiments to validate. We encourage readers to interpret Sec.~\ref{APP:analysis} accordingly.

\paragraph{Scope of the Neuroscientific Analysis.}
This paper is primarily an ML contribution: our goal is a substantially stronger decoding model together with a benchmark that makes fine-grained evaluation possible. We regard building models that can decode brain activity as a necessary step toward understanding how information is represented in the brain. The accompanying analysis is intended to demonstrate what the framework makes possible, not to settle neuroscientific questions.
A systematic, quantitative comparison of our attribution maps against established functional maps in the neuroimaging literature, including an assessment of whether any of the recovered patterns are genuinely novel, would constitute a substantial study on its own and is left to future work.

\section{Model Implementation Details}
\label{sec:additional_details}

\subsection{Architecture Details}
\label{sec:arch_details}

We follow the BIT model proposed in~\cite{beliy2026brainit}, using 128 voxel clusters. BIT consists of a Brain Tokenizer, which maps fMRI activations into 512-dimensional Brain Tokens via a single-head graph attention layer, and a Cross-Transformer Module with 5 self-attention blocks and 6 cross-attention blocks (8 heads each).
We modify BIT by increasing the hidden dimensionality from 512 to 768, matching the dimension of the soft prompts passed to the language model. We add 6 masked self-attention blocks (8 heads each), which allow the question to shape the extraction: the 32 learnable query tokens are concatenated with the question tokens, embedded using InstructBLIP's frozen Q-Former embedding layer, and these blocks let the query tokens attend to the question before each cross-attention step over the Brain Tokens. The mask prevents attention to padded positions in the question. The query token outputs are projected to 32 soft prompts of dimension 768.

We use \texttt{Salesforce/instructblip-flan-t5-xl}~\cite{dai2023instructblip}, which consists of a frozen EVA-CLIP ViT-g/14 image encoder (1408-dimensional features), a Q-Former module, and Flan-T5-XL~\cite{chung2022flan} as the frozen language model.

\subsection{Training Details}    
\label{sec:train_details}
We reserve 10\% of the training data for validation, 
which is used for hyperparameter selection and early stopping.

For training, we use the AdamW optimizer~\cite{loshchilov2019adamw} with a learning rate of $5 \times 10^{-4}$. Both stages are trained for 50 epochs, with a warmup period of 15 epochs in stage 1. The learning rate is reduced by a factor of 0.1 on validation plateau with patience 5, and the best checkpoint on validation data is saved.

In stage 2, Q-BIT LoRA uses $r = \alpha = 2$.


\subsection{Compute Resources}
\label{sec:compute_resources}
Brain-IT-VQA is trained on a single H200 GPU (joint-subject model). Stage 1 training completes in approximately 10 hours, and Stage 2 training requires approximately 18 hours, for a total of 28 hours per task/dataset. At inference, processing a single image with 20 queries takes 0.1s, demonstrating the practicality of the model for real-world applications.

\FloatBarrier
\vspace{0.3cm}
\section{NSD-VQA dataset generation}
\label{sec:NSD_VQA_APP}

\subsection{Human Evaluation}
\label{sec:human_evaluation}
To assess the quality of the automatically generated question-answer pairs, we conducted an independent human evaluation on a randomly sampled subset of NSD-VQA. Ten independent participants were presented with 50 randomly sampled images and five questions per image, balanced across the different question categories, resulting in 2,500 question-answer evaluations in total. For open-ended questions, participants were asked to select the correct answer from four multiple-choice options, with distractors sampled from other examples within the same semantic category. For every question, participants could additionally select an ``N/A'' option when they judged that the question was not applicable to the image or, for multiple-choice questions, that none of the provided answers was correct.

Across the evaluated cases, participants agreed with the generated annotations in 89.85\% of cases on average, while the ``N/A'' option was selected in 6.06\% of evaluations. The agreement rates across individual question categories are reported in Table~\ref{tab:human_eval}. Overall, these results indicate that the automatically generated question-answer pairs have high agreement with human judgments, supporting the use of NSD-VQA as a benchmark for fMRI-based visual question answering.

\vspace{1cm}
\begin{table}[h]
\centering
\caption{Human evaluation accuracy by question category. Accuracy is computed over valid responses, excluding ``N/A'' responses.}
\vspace{2mm}
\label{tab:human_eval}
\begin{tabular}{lc}
\toprule
\textbf{Category} & \textbf{Accuracy (\%)} \\
\midrule
Action & 85.71 \\
Animal & 96.63 \\
Color & 74.47 \\
Counting & 87.14 \\
Food & 95.12 \\
Holding & 92.86 \\
Location & 89.59 \\
Scene & 97.47 \\
Vehicle & 91.07 \\
Yes/No & 94.75 \\
\bottomrule
\end{tabular}
\end{table}

\newpage 
\vspace{-0.1cm}
\subsection{Dataset Statistics}
\begin{figure}[H]
    \centering
    \vspace{0.4cm}
    \includegraphics[
        width=\linewidth
    ]{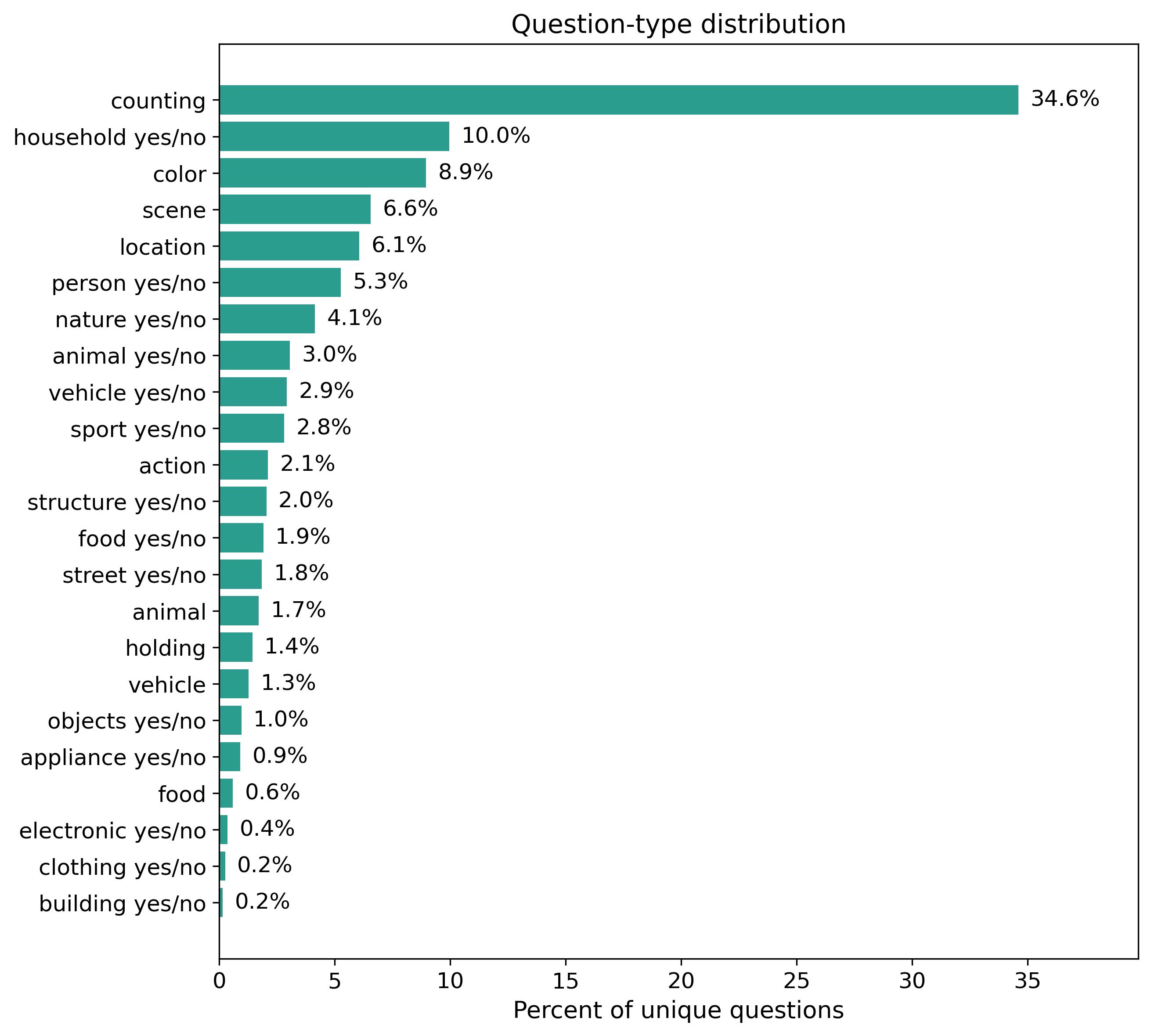}
    \vspace{-7mm}
    \caption{Distribution of question-answer pairs across NSD-VQA categories.}
    \label{fig:question_distribution}
    \vspace{-0.2cm}

\end{figure}

\vspace{0.1cm}
Fig.~\ref{fig:question_distribution} shows the distribution of question-answer pairs across NSD-VQA categories after filtering described in Sec.~\ref{sec:nsdvqa}. While NSD-VQA is organized around 20 semantic question categories, several categories are additionally separated into open-ended and binary (Y/N) variants for evaluation (e.g., animal vs.\ animal Y/N), resulting in the 23 displayed categories shown in Fig.~\ref{fig:question_distribution}.

\vspace{-3mm}


\subsection{Question Category Design and Scope}
NSD-VQA is not intended to provide an exhaustive taxonomy of neuroscientific questions. The Natural Scenes Dataset (NSD) was designed to capture a broad range of complex natural scenes rather than to probe any particular neuroscientific hypothesis. We therefore selected question categories based on what could be reliably annotated from the images and evaluated at sufficient scale. In particular, we retained distinctions that are explicitly observable in the images and occur frequently enough to support reliable evaluation, using a minimum support of 50 instances and filtering categories with highly imbalanced answer distributions.

The resulting categories cover several complementary aspects of visual understanding, including object identity and semantic category, counting, color and other attributes, actions and interactions, spatial position, and scene-level context. These distinctions are not claimed to be the uniquely most informative questions for neuroscience, nor to exhaust the space of visual processing. Rather, they provide a practical set of controlled dimensions that can be consistently grounded in the available NSD stimuli and decomposed for fine-grained analysis. The same annotation framework also allows the pipeline to construct additional question types when the corresponding visual evidence is present. This design choice follows the goal of NSD-VQA: to provide a reliable and interpretable benchmark for studying which aspects of visual and semantic information can be decoded from fMRI.

\subsection{Compute resources}        
\label{sec:nsdvqa_compute}

The NSD-VQA dataset construction pipeline, including annotation and verification using \texttt{Qwen3-VL-8B} and \texttt{Gemma-4-31B-it}, required approximately 30 GPU hours on a single H200 GPU. The generation of the full-sentence variant (NSD-VQA-FS) using Llama-3.2-3B required an additional 8 GPU hours. 

\subsection{Dataset Prompts}
\begin{figure}[h]
    \centering
    \includegraphics[
        width=\linewidth,
        height=\textheight,
        keepaspectratio
    ]{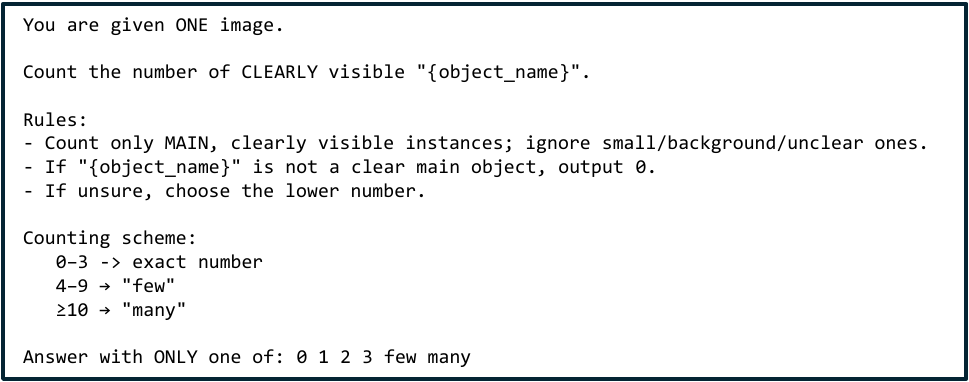}
    \caption{Counting verification prompt used for validating object counts during dataset construction.}
    \label{fig:counting_prompt}
\end{figure}

\begin{figure}[h]
    \centering
    \includegraphics[
        width=\linewidth,
        height=\textheight,
        keepaspectratio
    ]{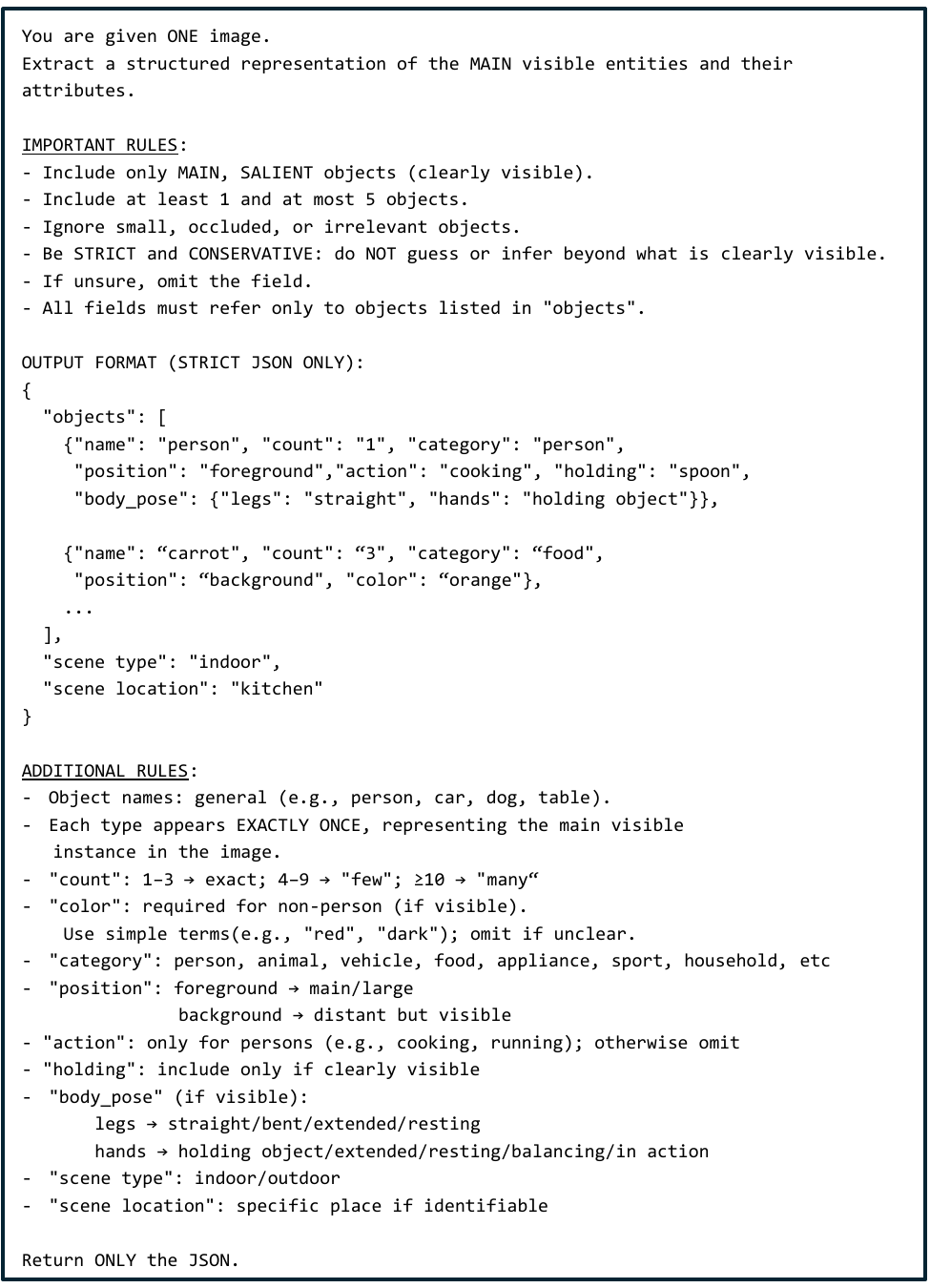}
    \caption{Structured annotation prompt used for extracting object- and scene-level attributes from NSD images.}
    \label{fig:annotation_prompt}
\end{figure}

Figs.~\ref{fig:annotation_prompt} and~\ref{fig:counting_prompt} show the prompts used during NSD-VQA construction. The annotation prompt is used to extract structured object- and scene-level information from NSD images using \texttt{Qwen3-VL-8B}\footnote{\url{https://huggingface.co/Qwen/Qwen3-VL-8B-Instruct}}, including object identity, attributes, actions, spatial position, and scene context. 

The counting verification prompt is used to validate object counts and object presence consistency during the filtering stage. Counts are independently estimated using both \texttt{Qwen3-VL-8B} and \texttt{Gemma-4-31B-it}\footnote{\url{https://huggingface.co/google/gemma-4-31B-it}}, and annotations are retained only when both models agree. Object presence consistency is additionally verified by requiring non-zero predicted counts across both models.

The generated structured annotations are subsequently converted into question-answer pairs using template-based generation rules conditioned on the available attributes and semantic categories.

\vspace{0.5cm}

\newpage
\section{Additional results}    
\label{sec:Add_results}

\subsection{Captioning Across Subjects}

\vspace{0.3cm}
\begin{table}[H]
\centering

\caption{Captioning performance of \textbf{Brain-IT-VQA} across subjects (1, 2, 5, and 7).} 
\label{tab:cap_metrics}
\begin{tabular}{lcccc}
\toprule
Metric & S1 & S2 & S5 & S7 \\
\midrule
CIDEr   & 0.683 & 0.646 & 0.722 & 0.598 \\
BLEU-1  & 68.11 & 66.26 & 69.11 & 65.39 \\
BLEU-2  & 49.30 & 47.54 & 50.51 & 46.00 \\
BLEU-3  & 35.08 & 33.48 & 36.12 & 32.00 \\
BLEU-4  & 24.81 & 23.37 & 25.71 & 22.19 \\
ROUGE   & 47.97 & 47.10 & 48.67 & 46.51 \\
METEOR  & 24.82 & 24.30 & 25.76 & 23.44 \\
SPICE   & 16.00 & 15.23 & 16.78 & 14.78 \\
\bottomrule
\end{tabular}
\vspace{2mm}
\end{table}

\vspace{0.3cm}
\subsection{VQA Results Across Subjects}
\label{sec:additional_vqa}

\vspace{0.3cm}
\begin{table}[H]
\centering
\label{tab:vqa_nsd_multisubject}
\small
\caption{Performance of \textbf{Brain-IT-VQA} across multiple datasets and subjects (1, 2, 5, and 7).}
\label{tab:vqa_nsd_multisubject}

\begin{tabular}{llcccc}
\toprule
\textbf{Dataset} & \textbf{Metric} 
& \textbf{S1} & \textbf{S2} & \textbf{S5} & \textbf{S7} \\
\midrule

\textbf{VQA-v2} 
& Accuracy $\uparrow$  
& 57.61 & 57.04 & 57.71 & 56.01 \\

\midrule
\multirow{8}{*}{\textbf{FSVQA}}
& VQA Acc. $\uparrow$ & 55.10 & 54.87 & 55.30 & 53.22 \\
& FSVQA Acc. $\uparrow$ & 52.00 & 51.73 & 51.80 & 49.53 \\
& BLEU-1 $\uparrow$ & 89.15 & 89.15 & 89.12 & 88.65 \\
& BLEU-2 $\uparrow$ & 86.14 & 86.12 & 86.06 & 85.50 \\
& BLEU-3 $\uparrow$ & 83.19 & 83.12 & 83.08 & 82.41 \\
& BLEU-4 $\uparrow$ & 80.17 & 80.03 & 79.99 & 79.16 \\
& METEOR $\uparrow$ & 52.40 & 52.28 & 52.38 & 51.69 \\
& CIDEr $\uparrow$ & 0.722 & 0.720 & 0.721 & 0.713 \\


\midrule

\multirow{4}{*}{\textbf{NSD-VQA}}
& Accuracy $\uparrow$
& 78.86 & 77.69 & 79.36 & 77.00 \\
& Acc (per-category) $\uparrow$
& 79.91 & 78.83 & 81.05 & 78.29 \\
& Acc (weighted) $\uparrow$
& 78.84 & 77.68 & 79.35 & 76.99 \\
& Acc (grouped) $\uparrow$
& 67.40 & 64.35 & 67.86 & 62.85 \\

\midrule

\multirow{7}{*}{\textbf{NSD-VQA-FS}}
& FSVQA Acc. $\uparrow$ 
& 78.31 & 78.24 & 79.36 & 76.80 \\
& BLEU-1 $\uparrow$   
& 94.47 & 94.42 & 94.75 & 94.04 \\
& BLEU-2 $\uparrow$   
& 92.87 & 92.79 & 93.23 & 92.30 \\
& BLEU-3 $\uparrow$   
& 91.30 & 91.21 & 91.76 & 90.61 \\
& BLEU-4 $\uparrow$   
& 89.52 & 89.40 & 90.09 & 88.67 \\
& METEOR $\uparrow$   
& 62.42 & 62.33 & 62.96 & 61.61 \\
& ROUGE-L $\uparrow$   
& 93.79 & 93.75 & 94.09 & 93.34 \\
& CIDEr $\uparrow$    
& 0.837 & 0.836 & 0.845 & 0.826 \\

\bottomrule
\end{tabular}
\vspace{2mm}

\end{table}

\FloatBarrier

\vspace{0.5cm}
\subsection{NSD-VQA Results per category}
\label{sec:additional_nsd_vqa_category}
\vspace{-0.5cm}

Table~\ref{tab:cat_all_subjects_split} reports NSD-VQA performance by category across subjects 1, 2, 5, and 7. Table~\ref{tab:cat_s1_models_split} compares Brain-IT-VQA against MindLLM on Subject 1 across categories. Statistical significance is evaluated using paired bootstrap testing with 10{,}000 bootstrap samples.

\begin{table}[H]
\centering
\caption{NSD-VQA performance by category across subjects (1, 2, 5 and 7).}
\label{tab:cat_all_subjects_split}
\small
\setlength{\tabcolsep}{5pt}
\renewcommand{\arraystretch}{1.05}

\begin{tabular}{lcccc|lcccc}
\toprule
\textbf{Category} & \textbf{S1} & \textbf{S2} & \textbf{S5} & \textbf{S7} &
\textbf{Category} & \textbf{S1} & \textbf{S2} & \textbf{S5} & \textbf{S7} \\
\midrule

action          & 61.99 & 56.07 & 59.81 & 55.14 &
household Y/N   & 90.66 & 89.94 & 90.53 & 89.09 \\

animal          & 67.82 & 62.07 & 67.43 & 60.92 &
location        & 56.93 & 52.16 & 57.14 & 53.03 \\

animal Y/N      & 92.26 & 89.68 & 93.76 & 88.82 &
nature Y/N      & 83.52 & 83.68 & 84.31 & 83.52 \\

appliance Y/N   & 96.40 & 91.37 & 96.40 & 92.81 &
objects Y/N     & 89.04 & 85.62 & 92.47 & 89.73 \\

building Y/N    & 73.91 & 86.96 & 82.61 & 86.96 &
person Y/N      & 93.27 & 93.52 & 94.14 & 92.52 \\

clothing Y/N    & 92.11 & 92.11 & 97.37 & 97.37 &
scene           & 95.00 & 93.30 & 93.80 & 93.00 \\

color           & 43.34 & 43.78 & 43.63 & 41.80 &
sport Y/N       & 95.33 & 91.36 & 93.46 & 93.69 \\

counting        & 79.67 & 79.19 & 80.67 & 78.25 &
street Y/N      & 85.71 & 88.57 & 87.86 & 89.29 \\

electronic Y/N  & 88.89 & 94.44 & 87.04 & 87.04 &
structure Y/N   & 88.75 & 87.46 & 90.03 & 89.07 \\

food            & 47.78 & 51.11 & 58.89 & 43.33 &
vehicle         & 70.98 & 66.32 & 70.98 & 60.62 \\

food Y/N        & 94.50 & 94.16 & 94.50 & 94.50 &
vehicle Y/N     & 89.01 & 90.58 & 92.15 & 87.89 \\

holding         & 61.01 & 49.54 & 55.05 & 52.29 &
                &       &       &       &       \\

\bottomrule
\end{tabular}
\end{table}

\vspace{0.5cm}
\begin{table}[H]
\centering
\caption{NSD-VQA performance by category for Subject 1. Mean and standard deviation are reported. Bold indicates statistically significant improvement over MindLLM under paired bootstrap testing ($p < 0.05$).}
\label{tab:cat_s1_models_split}
\small
\setlength{\tabcolsep}{4pt}
\renewcommand{\arraystretch}{1.05}

\begin{tabular}{lcc|lcc}
\toprule
\textbf{Category} & \textbf{MindLLM} & \textbf{BRAIN-IT VQA} &
\textbf{Category} & \textbf{MindLLM} & \textbf{BRAIN-IT VQA} \\
\midrule

action          & 49.53 $\pm$ 2.80 & \textbf{61.99 $\pm$ 2.74} &
household Y/N   & 85.86 $\pm$ 0.89 & \textbf{90.66 $\pm$ 0.75} \\

animal          & 51.34 $\pm$ 3.09 & \textbf{67.82 $\pm$ 2.90} &
location        & 41.13 $\pm$ 1.61 & \textbf{56.93 $\pm$ 1.63} \\

animal Y/N      & 87.31 $\pm$ 1.54 & \textbf{92.26 $\pm$ 1.24} &
nature Y/N      & 80.19 $\pm$ 1.58 & \textbf{83.52 $\pm$ 1.50} \\

appliance Y/N   & 90.65 $\pm$ 2.48 & \textbf{96.40 $\pm$ 1.59} &
objects Y/N     & 83.56 $\pm$ 3.09 & 89.04 $\pm$ 2.60 \\

building Y/N    & 73.91 $\pm$ 9.06 & 73.91 $\pm$ 9.26 &
person Y/N      & 90.27 $\pm$ 1.04 & \textbf{93.27 $\pm$ 0.88} \\

clothing Y/N    & 86.84 $\pm$ 5.53 & 92.11 $\pm$ 4.33 &
scene           & 91.70 $\pm$ 0.87 & \textbf{95.00 $\pm$ 0.69} \\

color           & 41.80 $\pm$ 1.33 & 43.34 $\pm$ 1.36 &
sport Y/N       & 93.22 $\pm$ 1.21 & 95.33 $\pm$ 1.03 \\

counting        & 75.86 $\pm$ 0.59 & \textbf{79.67 $\pm$ 0.55} &
street Y/N      & 84.29 $\pm$ 2.19 & 85.71 $\pm$ 2.12 \\

electronic Y/N  & 90.74 $\pm$ 3.94 & 88.89 $\pm$ 4.25 &
structure Y/N   & 84.89 $\pm$ 2.06 & 88.75 $\pm$ 1.80 \\

food            & 44.44 $\pm$ 5.26 & 47.78 $\pm$ 5.28 &
vehicle         & 54.40 $\pm$ 3.64 & \textbf{70.98 $\pm$ 3.25} \\

food Y/N        & 86.94 $\pm$ 1.96 & \textbf{94.50 $\pm$ 1.34} &
vehicle Y/N     & 86.77 $\pm$ 1.62 & 89.01 $\pm$ 1.48 \\

holding         & 46.33 $\pm$ 3.38 & \textbf{61.01 $\pm$ 3.29} &
                &                    &                    \\

\bottomrule
\end{tabular}

\vspace{2mm}
\end{table}

\FloatBarrier
\vspace{-0.2cm}
\subsection{NSD-VQA Question-Only Sanity Check}

Table~\ref{tab:VQA_baseline} reports a Question-Only Sanity Check obtained by evaluating the model without brain input, conditioning only on the textual question. Some binary (Y/N) categories achieve slightly above-chance performance because the evaluation split is not perfectly balanced between positive and negative answers, despite balancing being enforced at the full-dataset level. This allows the question-only baseline to partially exploit answer-frequency priors.

\vspace{-0.7cm}
\begin{table}
\centering
\vspace{-0.4cm}

\caption{NSD-VQA no fMRI baseline accuracy by category.}
\vspace{2mm}

\label{tab:VQA_baseline}
\begin{tabular}{lc|lc|lc}
\toprule
Category & Acc (\%) & Category & Acc (\%) & Category & Acc (\%) \\
\midrule
action & 14.02 & electronic Y/N & 46.30 & person Y/N & 54.99 \\
animal & 10.34 & food & 18.89 & scene & 70.90 \\
animal Y/N & 45.81 & food Y/N & 54.98 & sport Y/N & 43.93 \\
appliance Y/N & 59.71 & holding & 1.83 & street Y/N & 52.86 \\
building Y/N & 60.87 & household Y/N & 50.56 & structure Y/N & 50.80 \\
clothing Y/N & 52.63 & location & 13.53 & vehicle & 25.39 \\
color & 31.92 & nature Y/N & 45.80 & vehicle Y/N & 51.12 \\
counting & 44.76 & objects Y/N & 52.74 & & \\
\bottomrule
\end{tabular}
\vspace{2mm}
\end{table}

\FloatBarrier

\subsection{Failure Modes and Systematic Errors}
\label{sec:failure_modes}
This section expands the failure analysis summarized in Sec.~\ref{subsec:decodingperformance}. All results are pooled across subjects 1, 2 and 5, the subjects with the highest SNR and decoding performance, for which the error structure can be estimated most reliably.

\vspace{0.2cm}
\textbf{Structured confusions in open-ended categories.} When the model is wrong, it is wrong in a structured way: the predicted answer is typically close to the true answer in semantic space. Table~\ref{tab:confusions} lists, for each frequent true answer, the single most likely incorrect prediction and the fraction of instances receiving it. The dominant errors are within-category substitutions of visually and semantically similar objects (dog/cat, cow/horse, truck/car, standing/walking). This indicates that the limiting factor is discrimination among similar alternatives rather than detection, and accounts for the gap between the reliable binary accuracies and the lower open-ended accuracies in Table~\ref{tab:VQA_cat}. One interpretation is that similar objects may have similar fMRI representations. However, we cannot, rule out that the model also relies substantially on scene context as well.

\vspace{0.6cm}
\begin{table}[H]
\centering
\caption{\textbf{Systematic confusions in open-ended categories.} Most likely confusion per true answer, pooled across subjects 1, 2 and 5. $n$ is the number of questions with this true answer; rate is the fraction of those that received the given incorrect prediction. }
\label{tab:confusions}
\small
\begin{tabular}{llcll}
\toprule
\textbf{Category} & \textbf{True answer} & \textbf{$n$} & \textbf{Most likely error} & \textbf{Rate} \\
\midrule
Action  & standing       & 180  & walking     & 24\% \\
Action  & biking        & 24  & riding       & 25\% \\
Action  & eating & 36  & sitting & 44\% \\
Action  & snowboarding  & 40  & skiing       & 30\% \\
\midrule
Animal  & dog           & 104  & cat          & 19\% \\
Animal  & cat           & 104  & dog          & 24\% \\
Animal  & cow           & 80  & horse        & 19\% \\
Animal  & sheep         & 56  & horse        & 20\% \\
\midrule
Food    & donut         & 44  & cake       & 11\% \\
Food    & hot dog       & 16  & sandwich     & 50\% \\
Food    & broccoli        & 28  & salad        & 11\% \\
\midrule
Vehicle & bus           & 84  & car          & 20\% \\
Vehicle & truck       & 36  & car   &  47\% \\
Vehicle & train         & 148 & bus          &  9\% \\
\bottomrule
\end{tabular}
\vspace{0.4cm}
\end{table}

\textbf{Counting:} 
Counting reveals that the model captures presence and coarse quantity more reliably than exact enumeration. Performance is high for absence and single instances (88.1\% and 84.9\%), but drops markedly for \textit{two} (37.6\%), \textit{few} (37.6\%), and especially \textit{three} instances (18.7\%). Errors are predominantly downward, with \textit{two} most often predicted as \textit{one}, while \textit{three} is most often confused with \textit{one} or \textit{two}. Likewise, \textit{few} is frequently collapsed to \textit{one}. Thus, the model appears to recover rough cardinality and presence, but has difficulty distinguishing fine-grained quantities.

\vspace{0.2cm}
\textbf{Color}:
Color carries some signal in the fMRI response, but this information is strongly dependent on color type. Neutral colors are relatively recoverable, whereas vivid chromatic colors are not. Errors are also strongly biased toward neutral responses, suggesting that the model captures coarse color information but struggles to reliably distinguish specific chromatic categories. Color is known to be decodable from the visual cortex, but in these scenes it must be bound to a specific object within a complex visual context, which is substantially more difficult than decoding color information in isolation.

\newpage

\section{ablations}
\vspace{-0.1cm}
\label{sec:additional_ablations}
Our main results use the predicted-fMRI data augmentation described in
Sec.~\ref{sec:results}. We report results without it, and test whether its benefit is specific to
the encoder we use by repeating it with two alternatives: a CLIP-based
encoder~\cite{radford2021learning}, obtained by replacing the backbone of the encoder we use with
CLIP, and the Transformer brain encoder of Adeli et al.~\cite{adeli2026transformer}. Results are
reported in Tables~\ref{tab:aug_captioning} and~\ref{tab:aug_vqa}.

Even without augmentation, our model outperforms all prior methods on VQA-v2 and
FSVQA, and leads on all captioning metrics but CIDEr, where it is second to Brain-language
fusion~\cite{bosch2025brainlanguagefusionenablesinteractive}. The augmentation improves
performance with all three encoders, one differing in its vision backbone and the other in its
architecture. This indicates that image-to-fMRI encoders capture the structure of fMRI responses
well enough to generate training data that improves decoding, independently of the particular
encoder used.

\begin{table}[H]
\centering
\vspace{0.6cm}

\caption{\textbf{Effect of the predicted-fMRI augmentation on captioning}(subject 1). ``w/o external'' removes the predicted-fMRI augmentation entirely; the CLIP and Adeli variants replace the image-to-fMRI encoder used to generate the external data.}
\label{tab:aug_captioning}
\resizebox{\linewidth}{!}{%
\begin{tabular}{lcccccccc}
\toprule
\textbf{Method} & \textbf{BLEU-1} $\uparrow$ & \textbf{BLEU-2} $\uparrow$ & \textbf{BLEU-3} $\uparrow$ & \textbf{BLEU-4} $\uparrow$ & \textbf{METEOR} $\uparrow$ & \textbf{ROUGE} $\uparrow$ & \textbf{CIDEr} $\uparrow$ & \textbf{SPICE} $\uparrow$ \\
\midrule
MindLLM~\cite{mindllm2025} & 61.75 & 42.84 & 29.86 & 21.24 & 19.54 & 45.82 & 0.610 & -- \\
Brain-language fusion~\cite{bosch2025brainlanguagefusionenablesinteractive} & 62.10 & 43.70 & 29.80 & 20.40 & -- & 46.00 & 0.659 & -- \\
\midrule
Ours (w/o external)   & 66.62 & 47.56 & 33.12 & 22.86 & 23.85 & 47.03 & 0.633 & 14.60 \\
Ours (w/ CLIP enc.)   & 68.06 & \textbf{49.35} & \textbf{35.13} & \textbf{24.88} & 24.87 & 48.16 & \textbf{0.688} & 16.07 \\
Ours (w/ Adeli enc.)  & 67.72 & 49.03 & 34.84 & 24.45 & \textbf{25.00} & \textbf{48.25} & 0.684 & \textbf{16.42} \\
Ours (full)           & \textbf{68.11} & 49.30 & 35.08 & 24.81 & 24.82 & 47.97 & 0.683 & 16.00 \\
\bottomrule
\end{tabular}%
}
\vspace{0.3cm}
\end{table}


\begin{table}[H]
\centering
\vspace{0.4cm}
\caption{\textbf{Effect of the predicted-fMRI augmentation on VQA-v2 and FSVQA accuracy} (subject 1). We report our model trained without the predicted-fMRI augmentation, and with it using three image-to-fMRI encoders: a CLIP-based encoder, the Transformer brain encoder of Adeli et al., and the encoder used in our full model.}
\label{tab:aug_vqa}
\begin{tabular}{lcc}
\toprule
\textbf{Method} & \textbf{VQA-v2 Acc.} $\uparrow$ & \textbf{FSVQA Acc.} $\uparrow$ \\
\midrule
MindLLM~\cite{mindllm2025} & 52.14 & 43.00 \\
\midrule
Ours (w/o external)   & 55.14 & 47.93 \\
Ours (w/ CLIP enc.)   & 57.59 & 51.13 \\
Ours (w/ Adeli enc.)  & 56.72 & 50.53 \\
Ours (full)           & \textbf{57.61} & \textbf{52.00} \\
\bottomrule
\end{tabular}
\vspace{0.4cm}
\end{table}

To evaluate whether direct VQA decoding offers an advantage over image-based VQA, we compare Brain-IT VQA against InstructBLIP applied to Brain-IT reconstructed images. As shown in Table~\ref{tab:vqav2_comparison}, Brain-IT VQA outperforms the image-based approach, suggesting that decoding answers directly from brain activity is more effective than first reconstructing the image and then applying a VQA model. Nonetheless, Brain-IT (Images) remains a strong approach, surpassing all previous brain-to-VQA methods. For reference, we also report InstructBLIP applied to the ground truth images, which provides an upper bound on the performance attainable through this pipeline.

\begin{table}[h]
\centering
\vspace{0.7cm}

\small
\renewcommand{\arraystretch}{1.25}
\setlength{\tabcolsep}{9pt}
\caption{
\textbf{VQAv2 Accuracy Comparison.}
{\small \it Brain-IT VQA vs.\ InstructBLIP on reconstructed images vs.\ ground truth images (subject 1).}
}
\label{tab:vqav2_comparison}
\vspace{-0.01cm}
\begin{tabular}{@{}lc@{}}
\toprule
\textbf{Approach} & \textbf{VQAv2 Acc.} \\
\midrule
Brain-IT VQA        & 57.61 \\
Brain-IT (Images)   & 52.79 \\
Ground Truth Images & 72.28 \\
\bottomrule
\end{tabular}
\end{table}


\FloatBarrier

\clearpage

\section{Qualitative results}
\label{sec:Qualitative}

\begin{figure}[H]
    \centering
    \includegraphics[width=\textwidth]{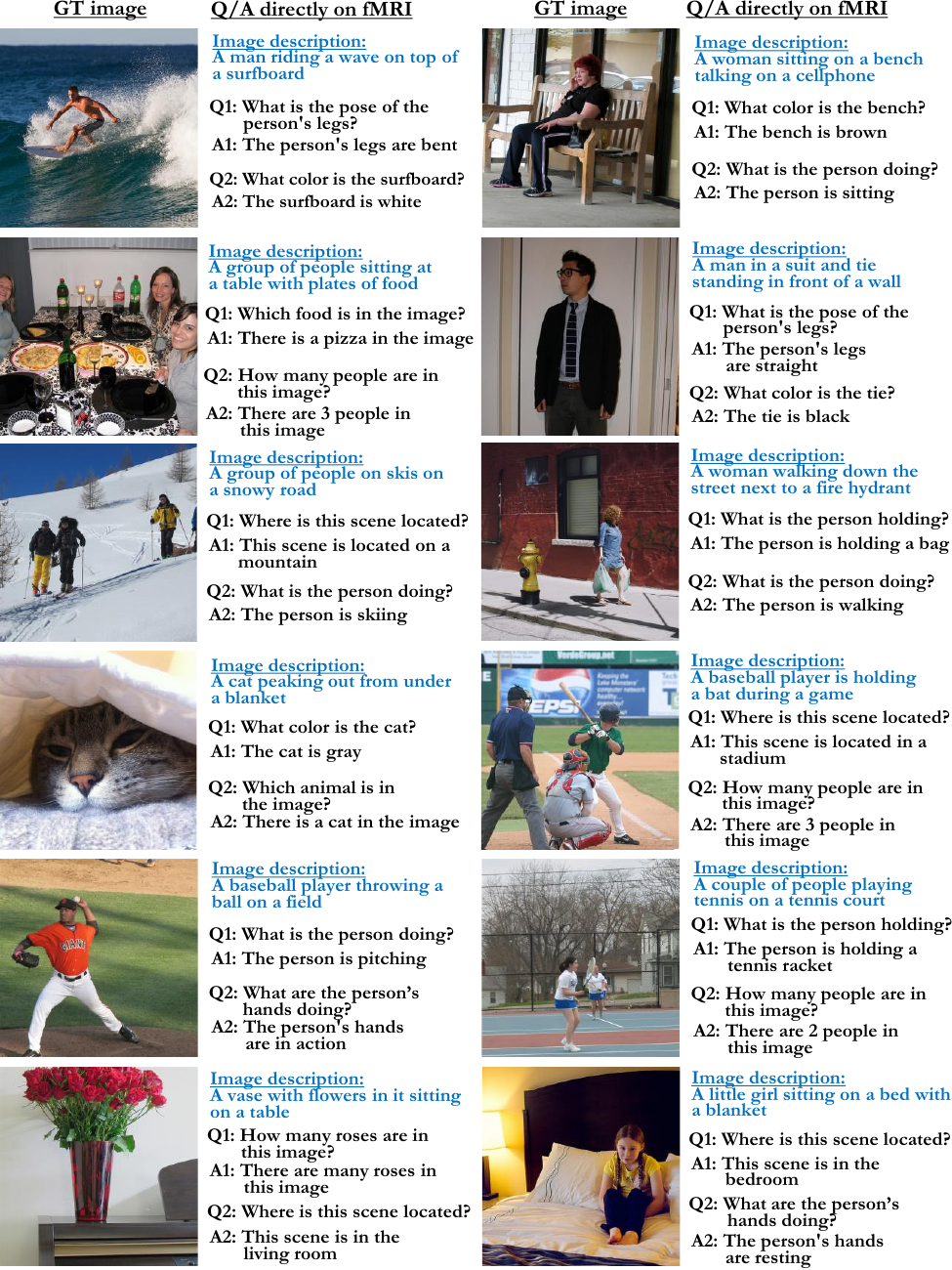}
    \caption{Additional qualitative results for \textit{Brain-IT-VQA}. The figure includes both generated image descriptions and answers to visual questions decoded directly from fMRI signals. Visual questions are drawn from the NSD-VQA-FS dataset.}
    \label{fig:additional_qual}
\end{figure}

\begin{figure}[h]
    \centering
    \includegraphics[width=\textwidth]{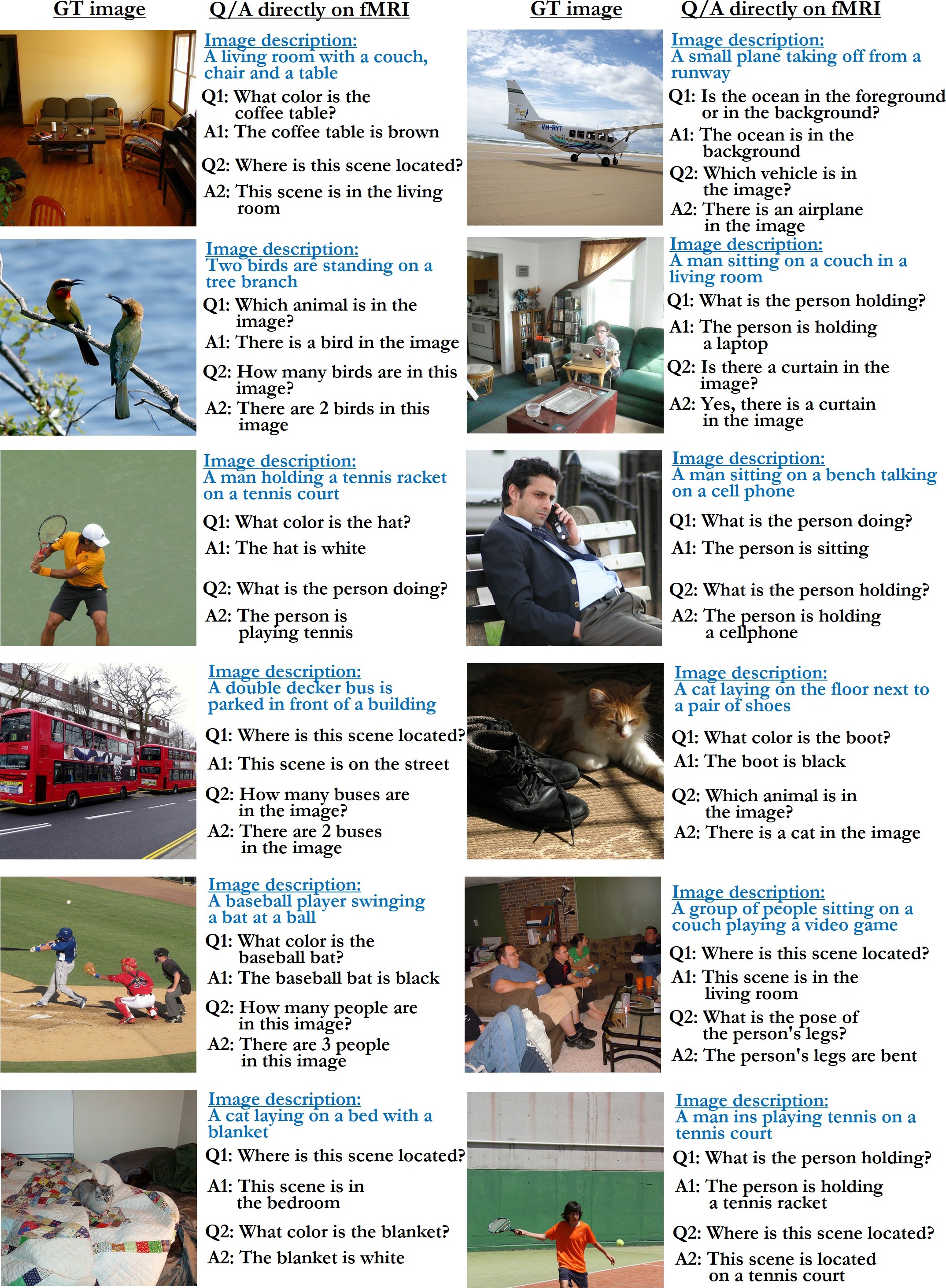}
    \caption{Additional qualitative results for \textit{Brain-IT-VQA}. The figure includes both generated image descriptions and answers to visual questions decoded directly from fMRI signals. Visual questions are drawn from the NSD-VQA-FS dataset.}
    \label{fig:additional_qual_2}
\end{figure}

Figs.~\ref{fig:additional_qual} and~\ref{fig:additional_qual_2} presents additional qualitative examples generated from fMRI signals using \textbf{Brain-IT-VQA}, including both image descriptions and answers to visual questions taken from NSD-VQA-FS.

\vspace{0.3cm}
\clearpage

\FloatBarrier

\newpage
\section{Decoding Contribution Analysis}
\label{APP:analysis}

\subsection{Masking Procedure}
\label{APP:masking}

Since Q-BIT processes fMRI inputs via a graph neural network, it naturally supports variable numbers of voxels. Masking is implemented by simply excluding the relevant voxels from the input graph. We use the Schaefer-1000 parcellation~\cite{schaefer2018local} for convenience, but any desired parcellation can be applied.

\subsection{Additional Brain Contribution Maps: Subject 1}
\label{APP:additional_sub1_maps}
\begin{figure}[h]
    \centering
    \includegraphics[width=1.2\textwidth]{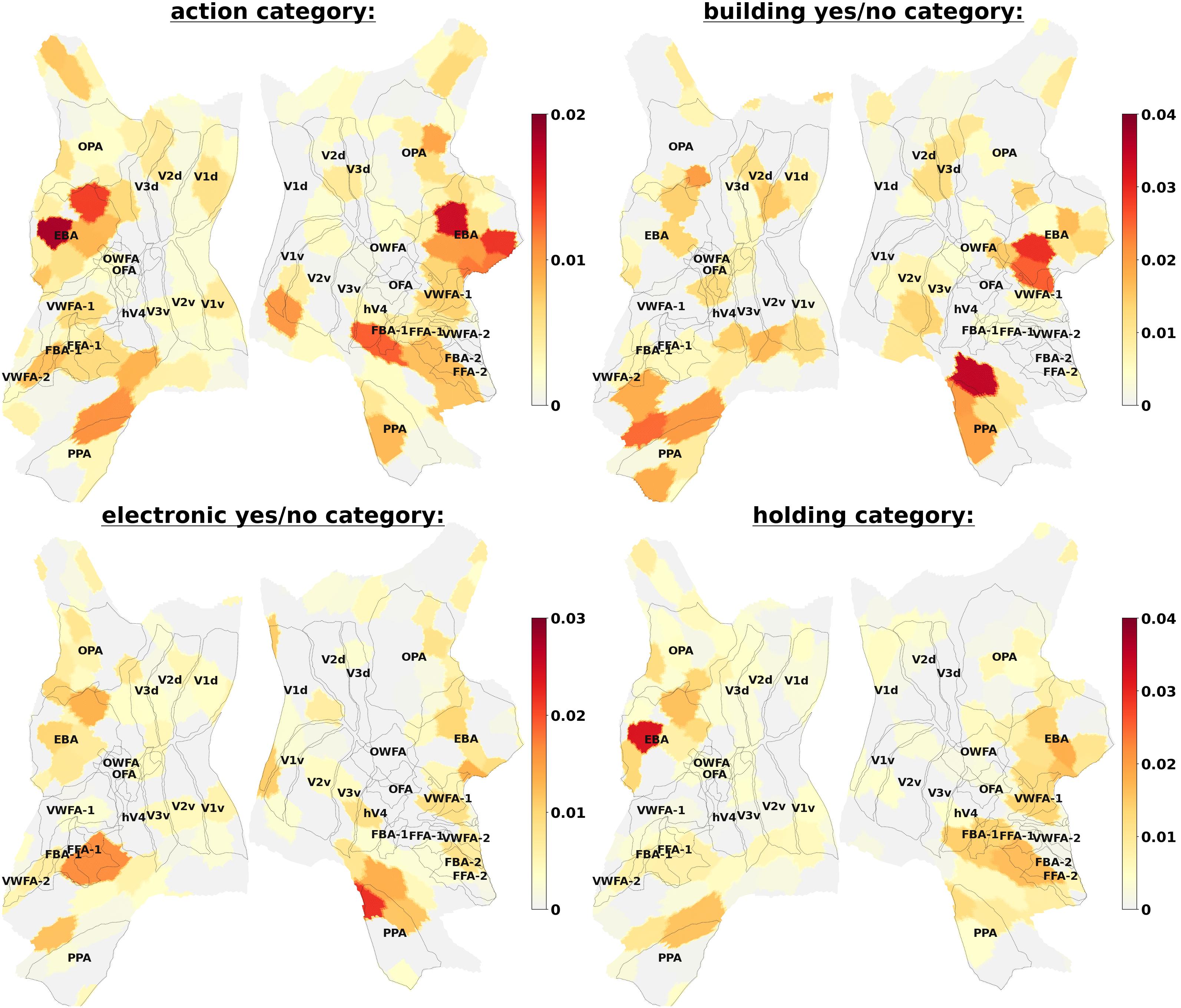}
    \caption{\textbf{Visualization of parcel marginal contributions across question categories, subject 1(a)}}
    \label{fig:additional_categories_1}
\end{figure}

\begin{figure}[h]
    \centering
    \includegraphics[width=1.2\textwidth]{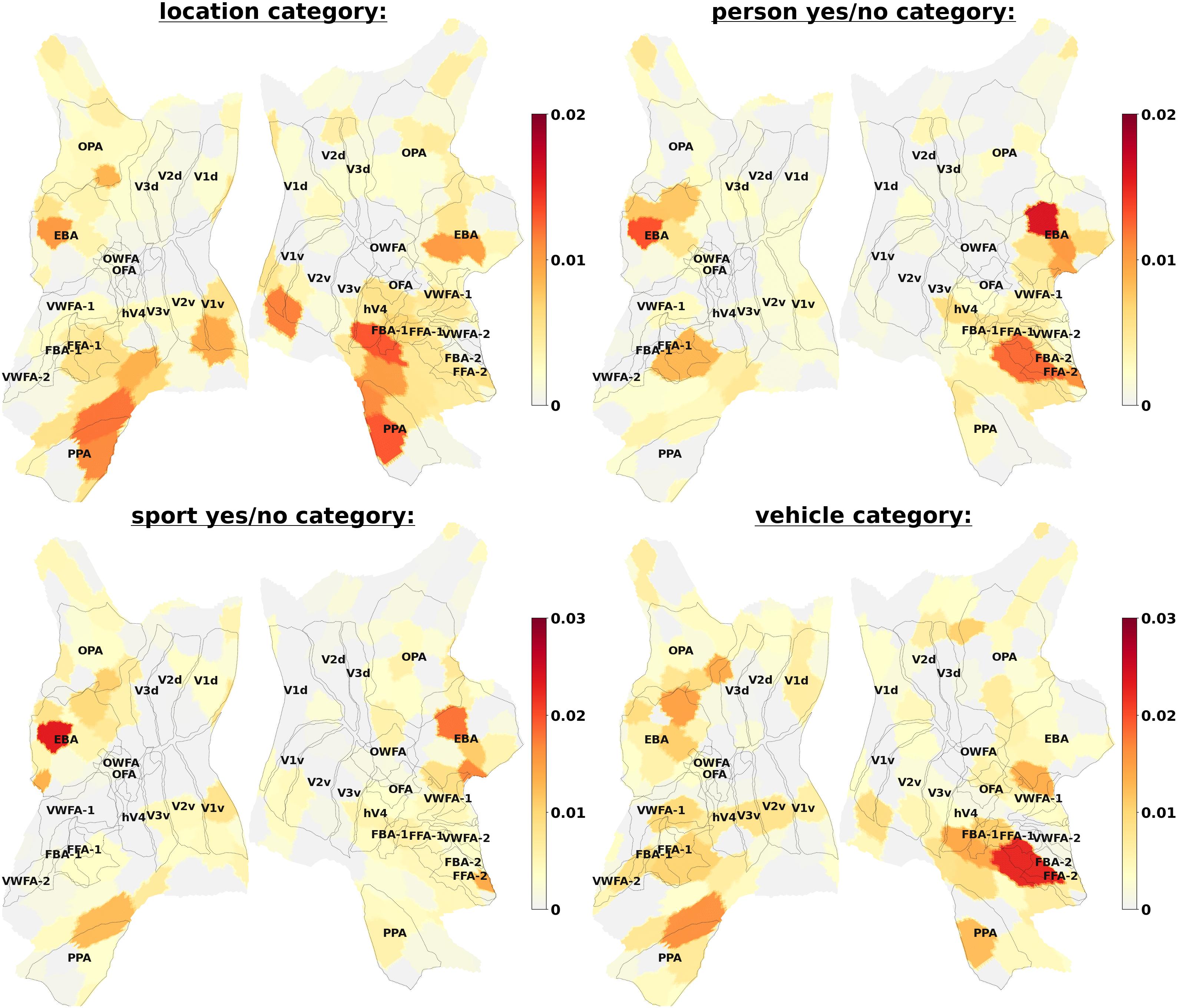}
    \caption{\textbf{Visualization of parcel marginal contributions across question categories, subject 1(b)}}
    \label{fig:additional_categories_2}
\end{figure}

Additional voxel-cluster contribution maps for subject 1 are shown in Figs.~\ref{fig:additional_categories_1}, ~\ref{fig:additional_categories_2}. They provide contribution
patterns across a broader set of question categories than those illustrated
in the main paper.
\label{APP:brain_plots}

\FloatBarrier
\clearpage

\subsection{BRAIN CONTRIBUTION MAPS: Additional Subjects}
\label{APP:brain_lots_additional_sub}
Additional contribution maps for subjects 2 and 5 are shown below. The maps exhibit broad consistency in the general location of contributing regions, while precise patterns vary across individuals.
\label{APP:brain_plots_sub}

\begin{figure}[h]
    \centering
    \includegraphics[width=1.2\textwidth]{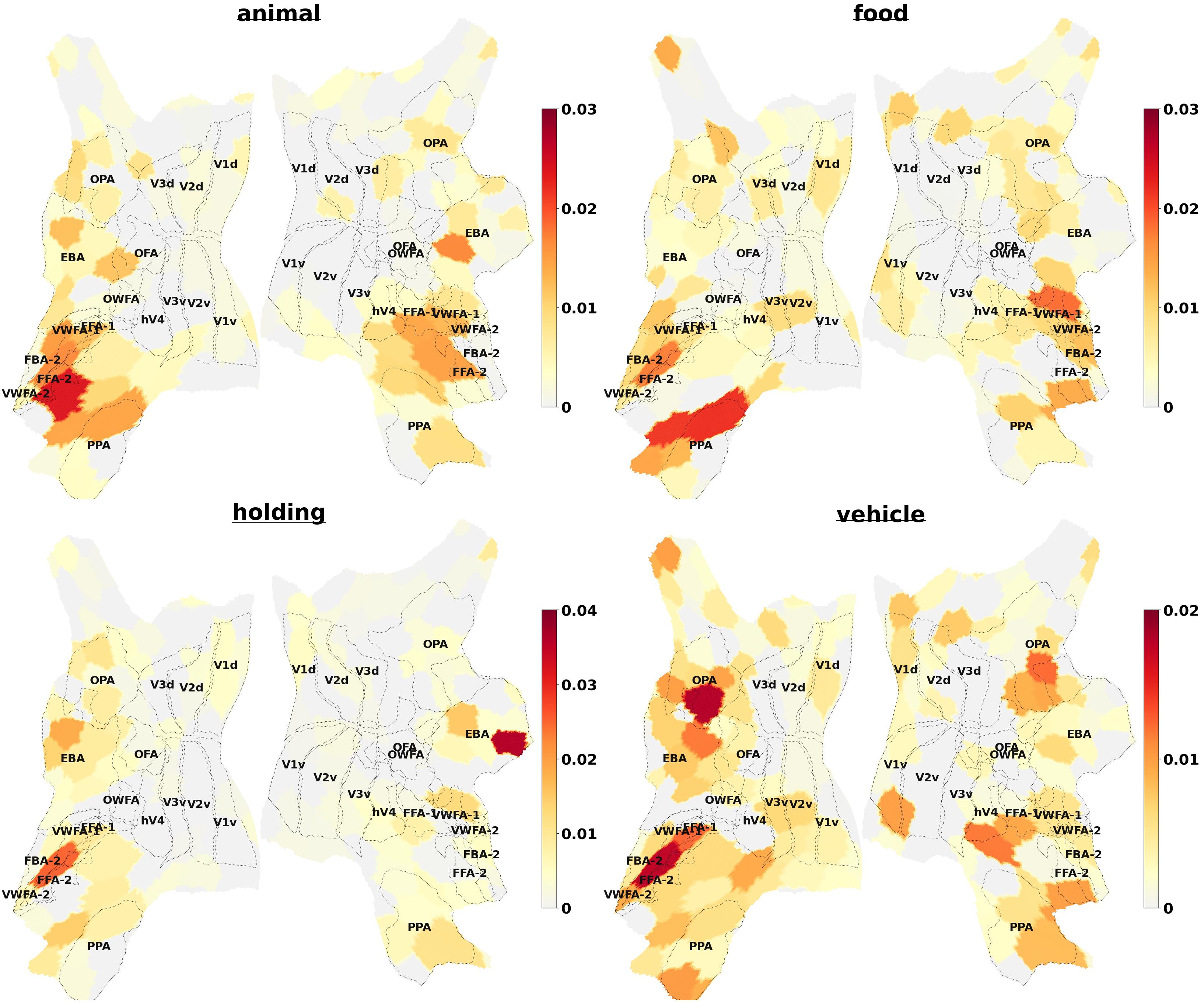}
    \caption{\textbf{Visualization of parcel marginal contributions across selected question categories, subject 2}}
    \label{fig:your_label}
\end{figure}

\begin{figure}[h]
    \centering
    \includegraphics[width=1.2\textwidth]{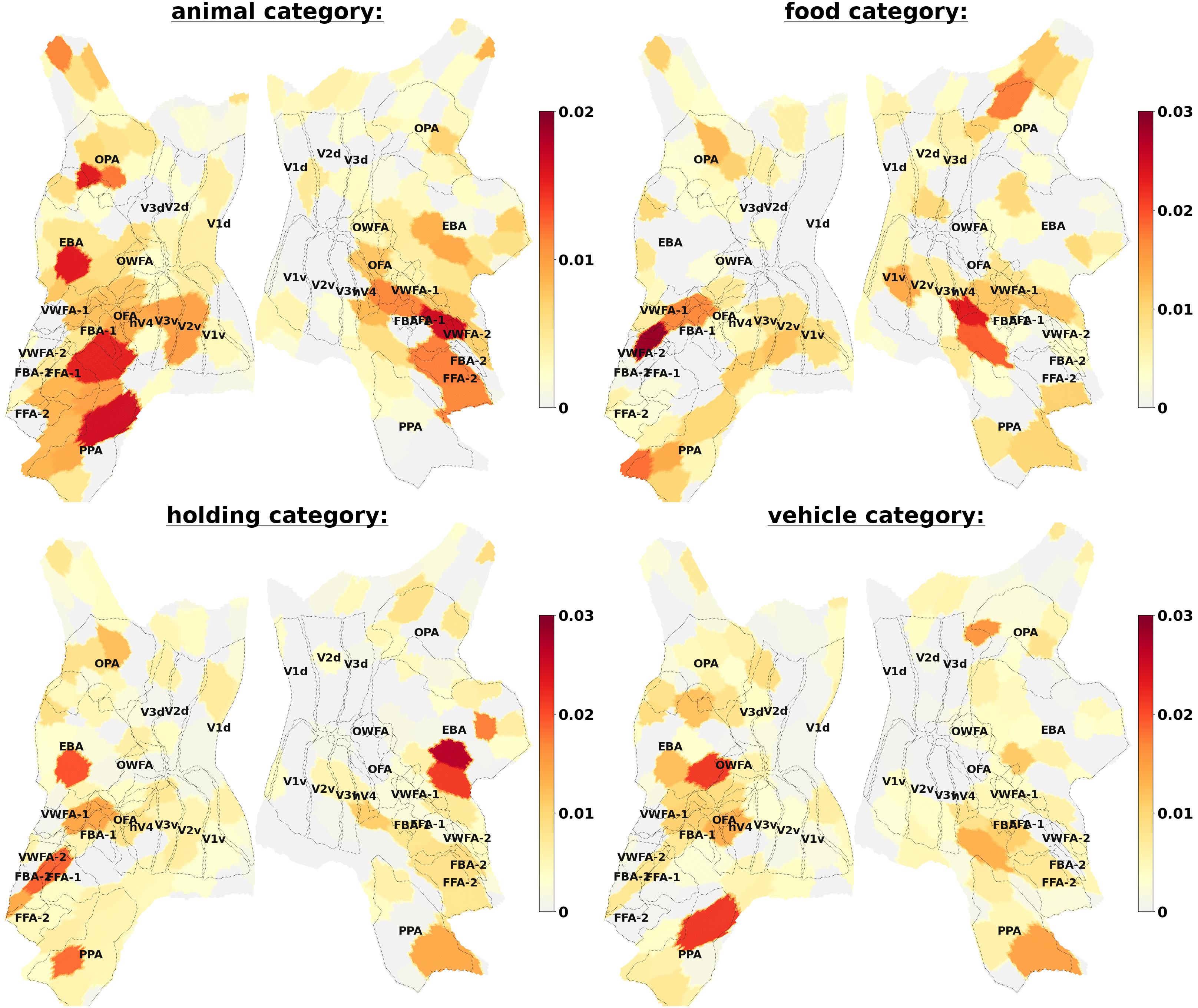}
    \caption{\textbf{Visualization of parcel marginal contributions across selected question categories, subject 5}}
    \label{fig:your_label}
\end{figure}

\FloatBarrier
\clearpage

\subsection{Extended Discussion of Contribution Patterns}
\label{APP:contribution_discussion}

The contribution maps reveal several recurring patterns across question categories. First, categories differ not only in the location of their contributing functional voxel groups, but also in the spatial distribution of these contributions. Human-object interaction categories such as \emph{holding} tend to exhibit relatively concentrated contributions,
whereas semantic categories such as \emph{food} show more distributed
patterns across ventral visual regions. Similar variation in spatial
distribution is visible across the additional categories shown in
App.~\ref{APP:additional_sub1_maps}.

Second, the parcel-level analysis reveals distinctions that can be obscured by coarser anatomical parcellations. In particular, different functional voxel groups within the same anatomical ROI can contribute differently to different question categories; for example, the maps show distinct contribution patterns for \emph{location} and \emph{action} within the broader PPA region. This suggests that aggregating activity at the ROI level can mask finer-grained category-dependent structure.

Third, the qualitative organization of the contribution maps is broadly
consistent across subjects, although the precise locations and relative
magnitudes of individual contributions vary. This variation is expected
given inter-subject differences in functional organization, while the
recurring category-dependent patterns provide evidence that the observed
structure is not restricted to a single subject (see App.~\ref{APP:brain_lots_additional_sub}).

These observations are intended as evidence of the exploratory value of the
proposed framework rather than as definitive functional localization claims.
The contribution estimates depend on the underlying decoding model and on
the available NSD stimuli; establishing correspondence with known
neuroimaging functional maps, or determining whether any observed patterns
are novel, would require a dedicated neuroscientific analysis beyond the
scope of this work.


\subsection{Functional ROI-Level Contributions}
Compared to the cluster-level analysis, ROIs represent a coarser partition of brain activity, 
with some regions containing thousands of voxels. 
EBA dominates contributions across most question categories, though this may be partly 
attributed to its disproportionately large number of voxels rather than functional specificity. 
Overall, the ROI-level analysis is less informative than the cluster-level results, 
motivating the use of finer-grained functional parcellations.
\label{APP:roi_plots}

\begin{figure}[H]
\centering
\vspace{0.3cm}
\includegraphics[width=\linewidth]{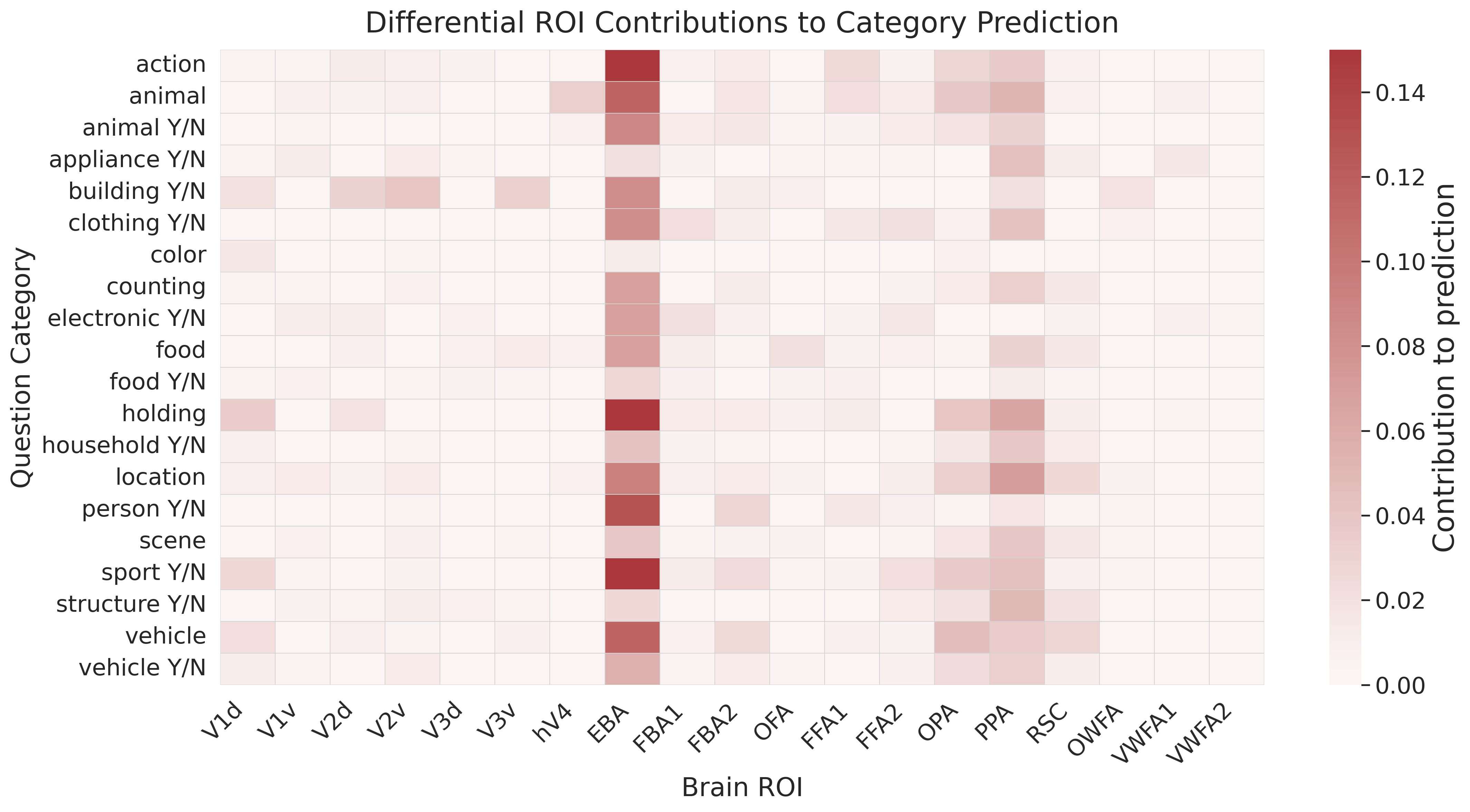}
\caption{Visualization of voxel-cluster contributions across question categories. Different clusters show varying levels of importance depending on the type of question (e.g., object, attribute, relation), highlighting how distinct brain regions support different aspects of visual and semantic processing.}
\label{fig:interpretability}
\end{figure}

\FloatBarrier

\end{document}